\documentclass[sn-mathphys]{sn-jnl}

\usepackage{graphicx}%
\usepackage{multirow}%
\usepackage{amsmath,amssymb,amsfonts}%
\usepackage{amsthm}%
\usepackage{mathrsfs}%
\usepackage[title]{appendix}%
\usepackage{xcolor}%
\usepackage{colortbl}%
\usepackage{textcomp}%
\usepackage{manyfoot}%
\usepackage{booktabs}%
\usepackage{algorithm}%
\usepackage{algorithmicx}%
\usepackage{algpseudocode}%
\usepackage{listings}%
\usepackage{arydshln}
\usepackage{subfig}

\definecolor{lightgreen}{HTML}{098842}  
\algnewcommand{\LineComment}[1]{         
  \State \textcolor{lightgreen}{\normalfont \(\triangleright\) \textit{#1}}
}

\begin{document}


\title[Domain Generalizable Continual Learning]{Domain Generalizable Continual Learning}


\author[1,3]{Hongwei Yan}\email{yanhw22@mails.tsinghua.edu.cn}

\author[1,3]{Guanglong Sun}\email{sgl23@mails.tsinghua.edu.cn}

\author[4]{Zhiqi Kang}\email{zhiqi.kang@inria.fr}

\author[1,3]{Yi Zhong}\email{zhongyithu@tsinghua.edu.cn}

\author*[2]{Liyuan Wang}\email{wly19@tsinghua.org.cn}

\affil[1]{School of Life Sciences, IDG/McGovern Institute for Brain Research, Tsinghua University, Beijing, China}

\affil[2]{Department of Psychological and Cognitive Sciences, Tsinghua University, Beijing, China}

\affil[3]{Tsinghua-Peking Center for Life Sciences, Beijing, China}

\affil[4]{Univ. Grenoble Alpes, Inria, CNRS, Grenoble INP, LJK, Grenoble, France}



\abstract{
To adapt effectively to dynamic real-world environments, intelligent systems must continually acquire new skills while generalizing them to diverse, unseen scenarios. Here, we introduce a novel and realistic setting named domain generalizable continual learning (DGCL): a model learns sequential tasks with each involving a single domain, aiming to perform well across all encountered tasks and domains. This setting poses unique challenges in acquiring, retaining, and leveraging both semantic- and domain-relevant information for robust generalization. Although state-of-the-art continual learning (CL) methods have employed pre-trained models (PTMs) to enhance task-specific generalization, they typically assume identical training and testing domains for each task and therefore perform poorly in DGCL. To this end, we propose adaptive Domain Transformation (DoT), an innovative PTMs-based approach tailored to DGCL. Inspired by the distributed-plus-hub theory of the human brain, DoT disentangles semantic- and domain-relevant information in representation learning, and adaptively transforms task representations across various domains for output alignment, ensuring balanced and generalized predictions. DoT serves as a plug-in strategy that greatly facilitates state-of-the-art CL baselines under both full parameter tuning and parameter-efficient tuning paradigms in DGCL, validated by extensive experiments. Also, DoT is shown to accumulate domain-generalizable knowledge from DGCL, and ensure resource efficiency with a lightweight implementation.
}


\maketitle

\section{Introduction}\label{sec.intro}

Human learning is characterized by the ability to learn only one task at a time in a single scenario, yet remember the underlying knowledge and generalize it well to other scenarios~\cite{kudithipudi2022biological,patterson2007you}. This remarkable adaptability underlies human intelligence in dynamic environments. Naturally, we expect artificial intelligence (AI) to adapt in a similar way. For example, a housekeeping robot must continuously keep up with emerging skill requirements, while adapting to varied user preferences and room layouts. Likewise, a healthcare diagnostic AI must continually incorporate medical imaging data from individual hospitals to improve disease identification, while generalizing to images from other hospitals with varying equipment and patient demographics. 
In this regard, we introduce a novel and realistic setting named domain generalizable continual learning (DGCL): a model learns sequential tasks with each involving a single domain, and is required to perform well across all encountered tasks and domains over time. 

The most relevant fields to DGCL are domain generalization (DG)~\cite{zhou2022domain,wang2022generalizing} and continual learning (CL)~\cite{wang2023comprehensive,parisi2019continual}, which have been traditionally studied in isolation and are not equipped to handle their integrated challenges.
DG aims to generalize from seen domains to unseen domains of individual tasks, assuming that the training data for all tasks is provided simultaneously. CL focuses on mitigating catastrophic forgetting when learning sequential tasks, assuming that the training and testing domains for each task are identical. In contrast, DGCL confronts both the continuity of learning tasks and the variability of testing domains, which poses unique challenges of acquiring, retaining, and leveraging semantic- and domain-relevant information for robust generalization. Although recent advances in CL have leveraged powerful pre-trained models (PTMs) to enhance task-specific generalization~\cite{wang2022learning_l2p,wang2022dualprompt,smith2023coda,wang2023hierarchical}, these methods are optimized for conventional CL settings and suffer significant performance degradation when applied to DGCL (see Sec.~\ref{sec.dgcl_exp}). 
In particular, many more advanced methods underperform relatively simple methods (i.e., L2P~\cite{wang2022learning_l2p} for parameter-efficient tuning and SLCA~\cite{zhang2023slca} for full parameter tuning) in DGCL, exposing severe limitations in both representation learning and output alignment.


\begin{figure}[t]
    \centering
    \vspace{-0.1cm}
    \includegraphics[width=0.80\textwidth]{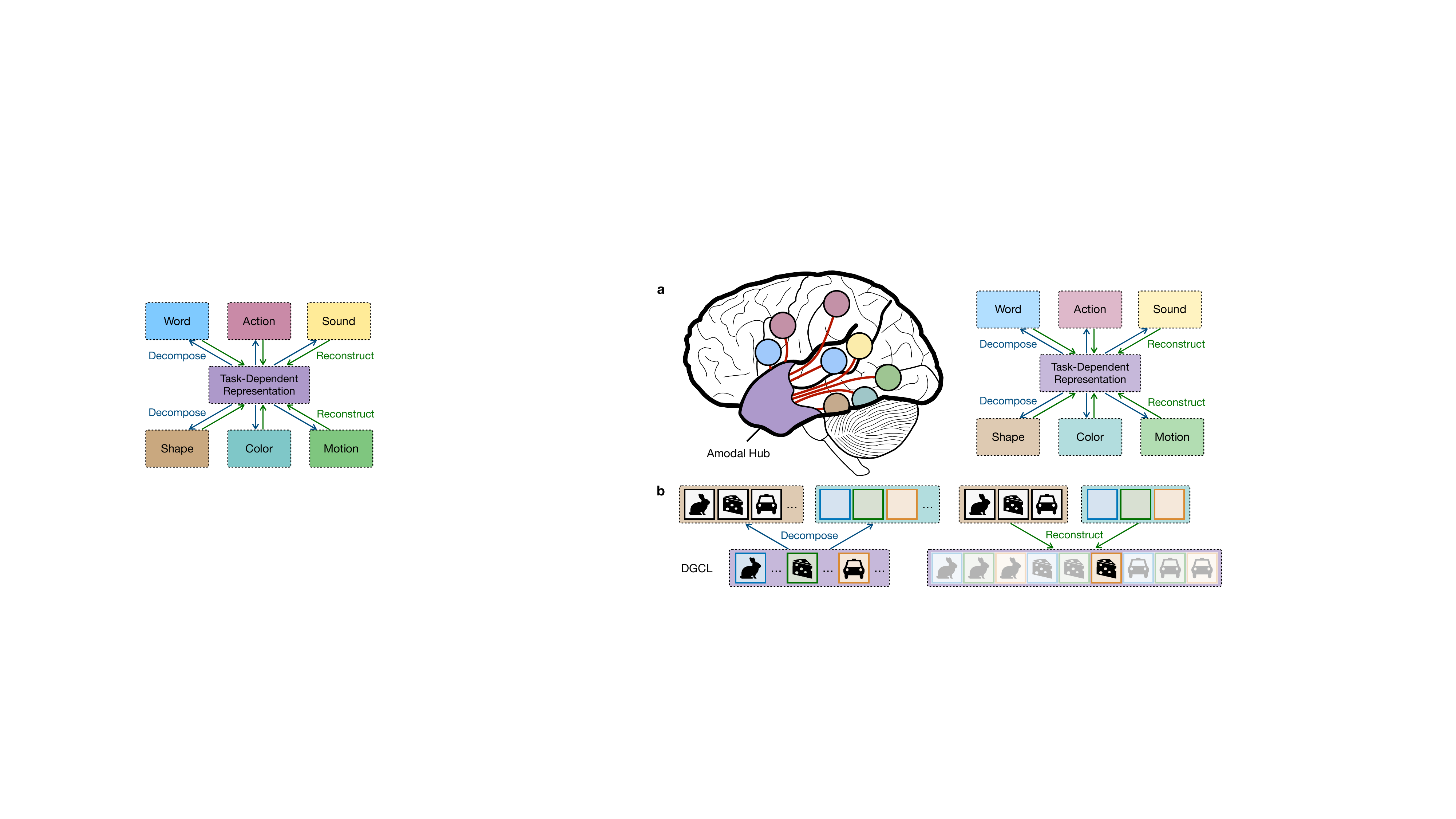}
\caption{Motivation of DGCL and DoT. \textbf{a}, The human brain naturally supports DGCL through distributed-plus-hub of memory attributes to reconstruct task-dependent representations~\cite{kudithipudi2022biological,patterson2007you}. \textbf{b}, Inspired by the human brain, we propose to address DGCL by coordinating semantic- and domain-relevant information in task-dependent representations. Best viewed in color.}
\vspace{-0.2cm}
\label{fig.distributed}
\end{figure}

To address these challenges, we draw inspiration from the human brain, which achieves robust DGCL-like capabilities though its organizing principles and memory consolidation mechanisms. Specifically, task-dependent experiences are consolidated into task-independent generalized knowledge after learning and are subsequently reconstructed in the light of such prior knowledge, as reflected in the neural representations with corresponding memory traces~\cite{lei2024reconstructing,frankland2005organization,barry2019remote}.
This process, encapsulated in the distributed-plus-hub theory, underpins cross-task and cross-domain generalization~\cite{patterson2007you} (see Fig.~\ref{fig.distributed}a).
With the neural inspiration, we propose adaptive Domain Transformation (DoT), an innovative PTMs-based approach tailored to DGCL. In representation learning, DoT obtains decoupled semantic- and domain-relevant information with the innate differentiation of layer-wise features in PTMs. The feature distributions corresponding to both types of information are effectively preserved, and combined flexibly with an attention-based transformation strategy. The transformed features encompass all encountered tasks and domains, collectively producing balanced and generalized predictions (see Fig.~\ref{fig.distributed}b).

We rigorously construct evaluation benchmarks for DGCL and demonstrate DoT's superiority with extensive experiments. DoT serves as a plug-in strategy that greatly facilitates state-of-the-art CL methods under both full parameter tuning and parameter-efficient tuning paradigms, allowing for significant advancements in DGCL. In addition, DoT is shown to accumulate domain-generalizable knowledge from DGCL, providing a more pronounced advantage of performing all seen tasks in completely unseen domains. DoT further ensures resource efficiency through a lightweight implementation, making it a practical choice for real-world applications.

Overall, our main contributions can be summarized as follows:
\begin{itemize}[nolistsep] 
\item We introduce DGCL, a novel and realistic setting that integrates the dual challenges of DG and CL in adapting to real-world complexity. 
\item We conduct an extensive empirical investigation of DGCL, revealing the severe limitations of cutting-edge advances in continual representation learning and continual output alignment.
\item We develop an innovative approach with reference to the human brain, which transforms semantic- and domain-relevant information in continually learned representations to achieve robust generalization.
\item Extensive experiments demonstrate the effectiveness of our approach, which significantly enhances state-of-the-art CL methods under both full parameter tuning and parameter-efficient tuning paradigms in the DGCL setting.
\end{itemize}

\section{Related Work}\label{sec.rw}

\textbf{Domain Generalization (DG)} aims to develop models capable of generalizing from one or more source domains to unseen target domains, without access to the latter during training \cite{zhou2022domain,wang2022generalizing}. Based on the composition of domain and label spaces, DG has been studied under several key settings. Multi-source DG leverages data from multiple source domains to learn domain-invariant representations~\cite{blanchard2011generalizing},
while single-source DG confronts the challenge of limited data diversity, often aligning closely with out-of-distribution (OOD) robustness research \cite{hendrycks2018benchmarking}.
Homogeneous DG assumes shared label spaces between source and target domains, whereas heterogeneous DG considers different label spaces between them~\cite{zhou2020learning}. 
Open-set DG~\cite{wang2023generalizable} adds previously unseen classes to seen classes in unseen target domains. Despite significant progress in capturing spatial-scale domain invariance (e.g., style normalization~\cite{jin2020style}), most existing methods lack the ability to encode semantic- and domain-relevant information across temporal scales. This limitation becomes critical when sequential tasks are tied to distinct source domains that evolve over time, rendering conventional DG methods inapplicable to the DGCL setting.

Several related settings extend the paradigm of DG, such as domain adaptation (DA)~\cite{wang2018deep}, test-time adaptation (TTA)~\cite{liang2024comprehensive}, and continual test-time adaptation (CTTA)~\cite{wang2022continual}, which leverage training or test data from target domains to enhance generalization. Specifically, DA employs sparsely labeled or unlabeled training data of target domains, TTA adapts to target domains using single or mini-batch test data, while CTTA extends TTA by accommodating dynamic target domains that evolve over time. In contrast, DGCL operates under the constraints of continual learning, where neither training nor test data from multiple tasks are accessible simultaneously. These realistic considerations make DGCL fundamentally distinct and necessitate novel approaches to address its unique challenges.

\begin{figure}[t]
    \centering
    \vspace{-0.1cm}
    \includegraphics[width=0.95\textwidth]{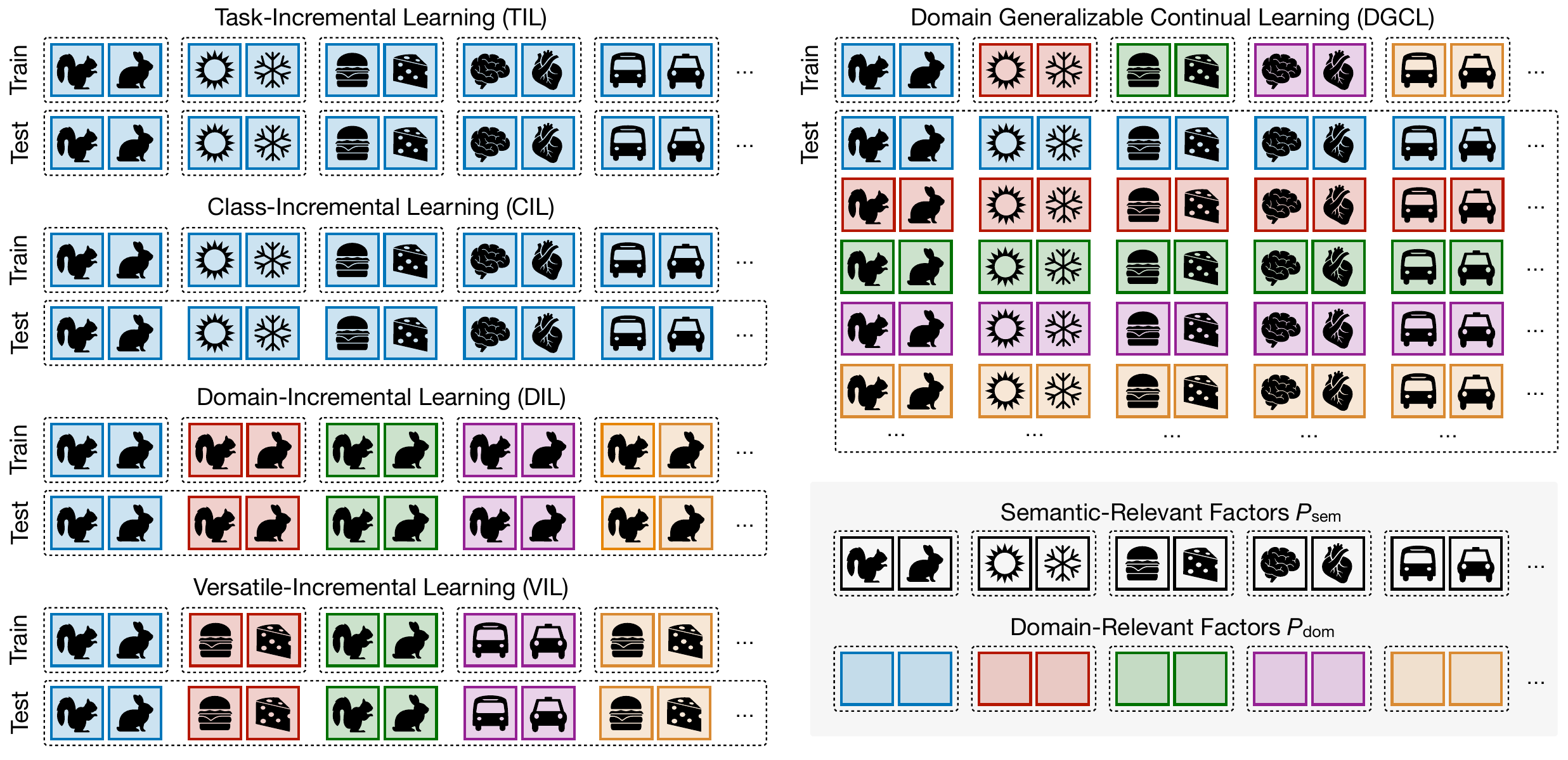}
\caption{Comparison of DGCL with representative CL settings, such as TIL, CIL, DIL, and VIL. DGCL is the only setting with training and test sets of each task belonging to different domains. }
\vspace{-0.2cm}
\label{fig.setting}
\end{figure}

\textbf{Continual Learning (CL)} aims to develop models to learn sequential tasks while mitigating catastrophic forgetting of previous knowledge~\cite{wang2023comprehensive,parisi2019continual,wu2024meta,wu2024mitigating}. Based on the composition of domain and label spaces, CL is typically categorized into task-incremental learning (TIL), class-incremental learning (CIL), and domain-incremental learning (DIL)~\cite{van2019three}. TIL and CIL involve disjoint label spaces, with TIL requiring task identity during testing, while DIL assumes a shared label space across tasks but varying domains.\footnote{For naming consistency, we refer to each training phase with a distinct data distribution as a ``task''.} Recent efforts have introduced more flexible paradigms, such as task-blurry incremental learning (TBIL)~\cite{moon2023online,kang2025advancing}, which allows overlapping label spaces in CIL, and versatile incremental learning (VIL)~\cite{park2025versatile}, which involves either new classes or new domains of old classes in each task. However, these settings inherently assume that the training and test data of each task share the same data distribution, failing to account for the broader challenges posed by DG (see Fig.~\ref{fig.setting}).

To address catastrophic forgetting, conventional CL methods are often categorized into regularization-based~\cite{wang2021afec,lyu2023overcoming}, replay-based~\cite{wang2021memory,yan2024orchestrate}, and architecture-based ones~\cite{wang2022coscl,wang2023incorporating}. Initially designed for training from scratch, these methods have shown limited effectiveness when applied to complex scenarios. Recent state-of-the-art CL methods transfer knowledge from PTMs to enhance generalization~\cite{zhang2023slca}. To avoid overwriting pre-trained parameters, these methods usually keep the pre-trained backbone frozen and adopt parameter-efficient tuning (PET) techniques for representation learning~\cite{wang2022learning_l2p,wang2022dualprompt,smith2023coda,wang2023hierarchical,wu2025sdlora}. When handling incremental tasks with distinct data distributions, they often employ task-shared or task-specific PET architectures, where the former requires gradual updates to mitigate catastrophic forgetting while the latter requires an additional query function to predict task identities~\cite{wang2024hide}. However, these methods are constrained by conventional CL settings, leaving their capabilities to address unseen domains of incremental tasks largely unexplored.

\section{Preliminaries}\label{sec.pre}

In this section, we first describe the problem formulation of our proposed DGCL setting, and then evaluate state-of-the-art CL methods under this new paradigm.

\subsection{Problem Formulation}\label{sec.problem}

DGCL requires a model to learn a sequence of tasks from their individual training sets, each associated with a potentially distinct domain (i.e., the domains may be identical or different from task to task), while generalizing to all encountered tasks and domains at test time. Let $\mathcal{X}$ and $\mathcal{Y}$ denote the global input and output spaces, respectively. For each task $t \in \{1, \dots, T\}$, the input space $\mathcal{X}_t \subseteq \mathcal{X}$ comprises both semantic- and domain-relevant information. Specifically, the input data is drawn from a joint distribution $P_X^{(t)} = P_{\text{dom}}^{(t)} \circ P_{\text{sem}}^{(t)}$, where $P_{\text{dom}}^{(t)}$ denotes the domain-relevant factor and $P_{\text{sem}}^{(t)}$ denotes the semantic-relevant factor (see examples in Fig.~\ref{fig.setting}). Each task is further characterized by a semantic distribution $P_Y^{(t)}$ over its output space $\mathcal{Y}_t \subseteq \mathcal{Y}$.

The training set for each task $t$ is defined as $\mathcal{D}_t = \{(\boldsymbol{x}_{t,n}, y_{t,n})\}_{n=1}^{N_t}$, where $\boldsymbol{x}_{t,n} \sim P_X^{(t)} = P_{\text{dom}}^{(t)} \circ P_{\text{sem}}^{(t)}$, $y_{t,n} \sim P_Y^{(t)}$, and $N_t$ is the total number of training samples. The input data of task $t$ originates from a particular source domain $d_t$ with domain-relevant factor $P_{\text{dom}}^{(t)}$, and its semantic-relevant factor $P_{\text{sem}}^{(t)}$ aligns with the task-specific semantic distribution $P_Y^{(t)}$. Notably, the input domains may differ across tasks, i.e.,  $\exists i \neq j$, $ P_{\text{dom}}^{(i)} \neq P_{\text{dom}}^{(j)}$ and $d_i\neq d_j$; while the semantic-relevant factors remain disjoint as in regular CL settings, i.e., $\forall i\neq j, P_{\text{sem}}^{(i)}\neq P_{\text{sem}}^{(j)}$ and $P_{Y}^{(i)}\neq P_{Y}^{(j)}$.
At the testing phase of $T$ tasks, the domain-relevant factors of all tasks $1, \dots, T$ are combined into a unified one $\bigcup_{i=1}^T P_{\text{dom}}^{(i)}$ from all domains $d_i$, while the semantic distribution of each task $t$ remains aligned with its original $P_{\text{sem}}^{(t)}$ and $P_Y^{(t)}$. Consequently, the test data for task $t$ is sampled as $\boldsymbol{x} \sim \bigcup_{i=1}^T P_{\text{dom}}^{(i)} \circ P_{\text{sem}}^{(t)}$ while its ground-truth label $y \sim P_Y^{(t)}$ preserves the original distribution. 

Consider a neural network model comprising a backbone $f_\theta(\cdot)$ with parameters $\theta$ and an output layer $h_\psi(\cdot)$ with parameters $\psi$. The objective is to learn a projection from $\bigcup_{t=1}^T \mathcal{X}_t$ to $\bigcup_{t=1}^T \mathcal{Y}_t$ by minimizing empirical errors across sequentially observed tasks and domains, so that the model can correctly predict the label $\hat{y} = h_\psi(f_\theta(\boldsymbol{x}))$ for unseen test data $\boldsymbol{x}$. 
However, learning such a desirable projection is highly nontrivial in DGCL: while the training data for each task is confined to a single source domain, the test data for ultimate evaluation spans all encountered tasks and domains. Distinct from TIL, CIL, DIL, and VIL paradigms, DGCL uniquely combines CL with task-specific DG, addressing the integrated challenges of mitigating catastrophic forgetting, managing disjoint label spaces, and generalizing across different domains (see Fig.~\ref{fig.setting}). These challenges are reinforced by the inherent CL constraint of task-specific data isolation - historical training samples stay strictly \emph{inaccessible} during the incremental learning process.

\subsection{CL with PTMs}\label{sec.pet_ptms}

State-of-the-art CL methods leverage advanced PTMs to enhance task-specific generalization. The pre-trained backbone $f_\theta (\cdot)$ often employs transformer-based architectures~\cite{vaswani2017attention}, such as the Vision Transformer (ViT) \cite{dosovitskiy2020image_vit}. ViT processes input data via consecutive multi-head self-attention layers (assuming a total of $L$ layers), producing sequence-like token embeddings $\boldsymbol{h}^{(L)} \in \mathbb{R}^{s \times m}$, where $s$ is the sequence length and $m$ is the embedding dimension. For each layer $l \leq L$, relationships within the input $\boldsymbol{h}^{(l-1)}\in \mathbb{R}^{s \times m}$ are modeled by computing attention scores for queries $\boldsymbol{h}_Q^{(l)}$, keys $\boldsymbol{h}_K^{(l)}$, and values $\boldsymbol{h}_V^{(l)}$ using learnable projection matrices $\boldsymbol{W}_Q^{(l)}$, $\boldsymbol{W}_K^{(l)}$, and $\boldsymbol{W}_V^{(l)}$, respectively, resulting in the layer-wise output $\boldsymbol{h}^{(l)}$ that captures both global and local context.  

When updating the entire parameter set $\theta$, i.e., full parameter tuning, $f_\theta(\cdot)$ can effectively adapt its output representations to specific tasks, but this often leads to severe forgetting of the pre-trained knowledge. A common alternative is to keep $f_\theta(\cdot)$ frozen and employ PET techniques for lightweight modifications. For example, Prompt Tuning (ProT) and Prefix Tuning (PreT) both add a set of learnable prompts to the layer-wise input: ProT prepends prompts to keys, queries, and values, whereas PreT divides them into task-specific prefixes for keys and values. Adapter Tuning (AdaT)~\cite{rebuffi2017learning} inserts lightweight neural modules between transformer layers, while Low-Rank Adaptation (LoRA)~\cite{hu2021lora} approximates the updates to specific projection matrices (e.g., $\boldsymbol{W}_Q^{(l)}$ and $\boldsymbol{W}_V^{(l)}$) via low-rank decomposition for efficient adaptation.

In general, recent efforts in CL with PTMs focus on two aspects for sequential tasks: continual representation learning for $f_\theta (\cdot)$ and continual output alignment for $h_\psi(\cdot)$, as summarized below. Their effectiveness in DGCL will be examined in Sec.~\ref{sec.dgcl_exp}.

\textbf{Representation Learning:} Existing methods have developed a variety of PET architectures, including task-specific parameters, which avoid inter-task interference but require additional inference of task identities; task-shared parameters, which avoid additional inference of task identities but need to address catastrophic forgetting during parameter updates; or a combination of both. 
As a pioneering work, L2P~\cite{wang2022learning_l2p} constructs a pool of prompts, each associated with a learnable key, and searches the most relevant prompts based on the cosine similarity between uninstructed representations and these keys.
DualPrompt~\cite{wang2022dualprompt} incorporates either task-specific or task-shared prompts in different layers, while S-Prompt~\cite{wang2022sprompts} and HiDe~\cite{wang2023hierarchical,wang2024hide} employ only task-specific prompts. CODA-Prompt~\cite{smith2023coda} performs a weighted summation over an expandable prompt pool, implementing orthogonal regularization to reduce inter-task interference. LAE~\cite{gao2023unified} and RanPAC~\cite{mcdonnell2023ranpac} adopt task-shared parameters, which are updated at a reduced pace or even frozen during CL.

\textbf{Output Alignment:} Many recent methods approximate and replay pre-trained representations of previous tasks to rectify potential bias in $h_\psi (\cdot)$. For example, SLCA~\cite{zhang2023slca,zhang2024slca++} performs full parameter tuning with a reduced learning rate, preserving pre-trained representations of each class with dedicated mean and covariance matrices. RanPAC~\cite{mcdonnell2023ranpac} expands pre-trained representations into a high-dimensional space, maintaining the inherent structure with a shared covariance matrix. HiDe~\cite{wang2023hierarchical,wang2024hide} evaluates various strategies for preserving pre-trained representations, such as using randomly selected prototypes, multiple centroids, mean-variance, and mean-covariance. ICON~\cite{park2025versatile} performs adaptive classifier expansion and cluster-based regularization to accommodate either new tasks or new domains of old tasks.

\begin{figure}[t]
    \centering
    \vspace{-0.1cm}
    \includegraphics[width=1.00\textwidth]{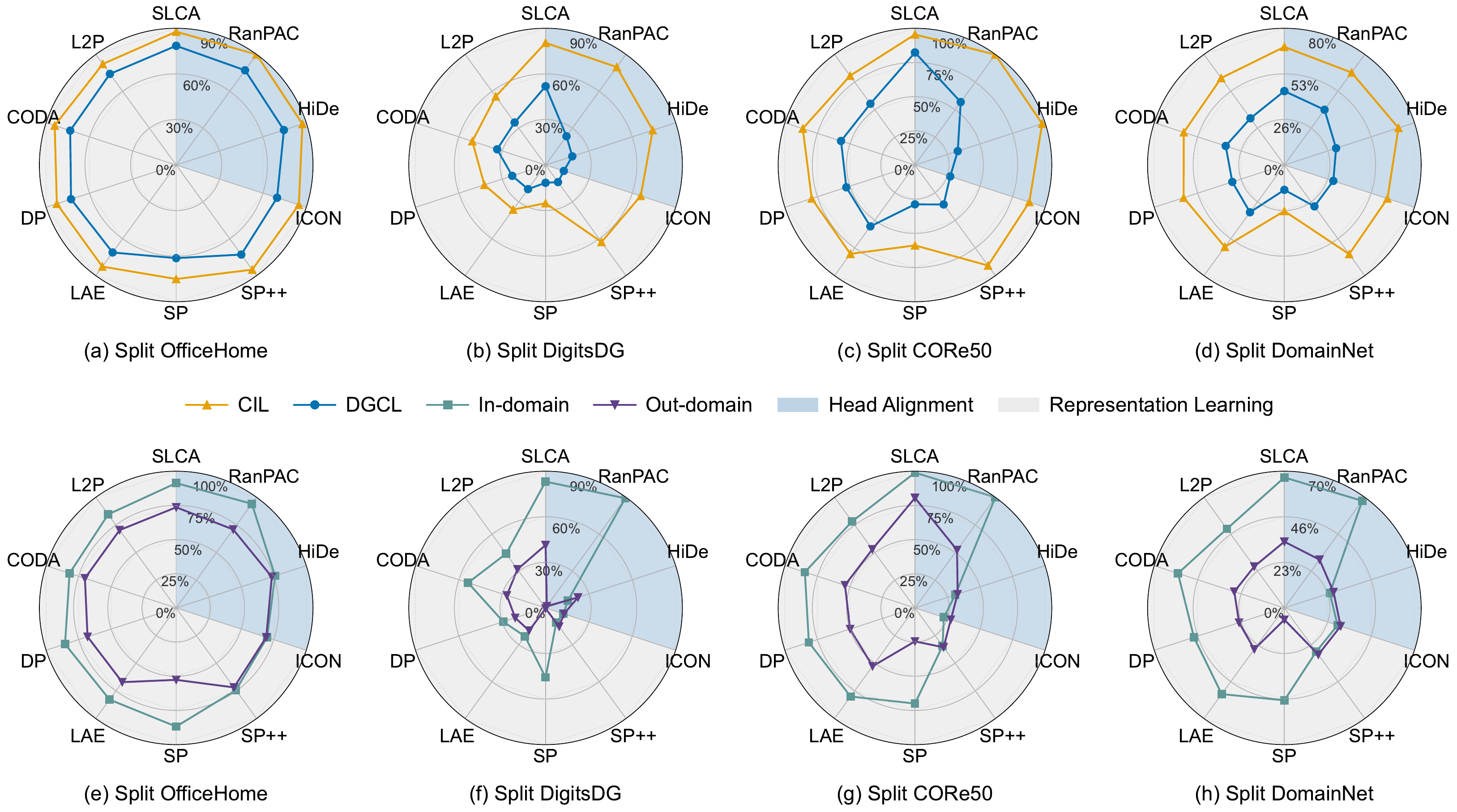}
    \vspace{-0.1cm}
\caption{Empirical investigation of DGCL with state-of-the-art CL baselines. \textbf{a}-\textbf{d}, We report the average accuracy of all encountered tasks and domains under CIL and DGCL settings (i.e., average all-domain accuracy $A_{\text{all}}$). \textbf{e}-\textbf{h}, We further report the average accuracy of tasks with seen domains (i.e., average in-domain accuracy $A_{\text{in}}$) and tasks with unseen domains (i.e., average out-domain accuracy $A_{\text{out}}$). Please refer to Sec.~\ref{sec.exp_setup} for implementation details. Methods in light gray focus on instructing $f_\theta (\cdot)$, while those covered in light blue focus on rectifying $h_\psi (\cdot)$ with or without instructing $f_\theta (\cdot)$. DP, DualPrompt. SP, S-Prompt. CODA, CODA-Prompt.}
\vspace{-0.2cm}
\label{fig.CIL_DGCL}
\end{figure}

\section{Empirical Investigation of DGCL}\label{sec.dgcl_exp}

We perform an extensive empirical investigation to analyze the particular challenges of DGCL. We construct sequential tasks with well-established class recognition datasets composed of distinct domains, including Office-Home~\cite{venkateswara2017deep}, DigitsDG~\cite{zhou2020learning}, CORe50~\cite{lomonaco2017core50}, and DomainNet~\cite{peng2019moment}. Experiments are conducted in both CIL and DGCL settings. Specifically, we first split class labels into disjoint task-specific subsets. In CIL, data from all domains is available for each task. Nevertheless, in DGCL, only one randomly assigned domain is available for each task, restricting its training data to that source domain exclusively (see Sec.~\ref{sec.exp_setup} for implementation details). We consider a wide range of state-of-the-art baselines described in Sec.~\ref{sec.pet_ptms}. Most of them are designed for CIL, except S-Prompt for DIL and ICON for VIL.

\textbf{Overall Performance}: We compare the average accuracy of classes across all encountered tasks and domains under both CIL and DGCL settings, as shown in Fig.~\ref{fig.CIL_DGCL} \textbf{a}-\textbf{d}. State-of-the-art baselines exhibit significant performance degradation in DGCL, primarily due to the limited availability of domain-relevant information underlying the associated semantic-relevant information in sequential tasks. 
Notably, L2P, the first PET-based CL method, is inferior to other more advanced methods in CIL but achieves almost the highest overall performance in DGCL. SLCA, a simple baseline that performs full parameter tuning with a reduced learning rate and performs representation recovery for output alignment, achieves overwhelmingly better performance than all PET-based CL methods in DGCL. S-Prompt and ICON, despite their specific design for incremental domains, cannot outperform other CIL methods. These results underscore that cutting-edge advances in CL are less effective in resolving DGCL.

\textbf{In- vs Out-Domain}: We further compare the average accuracy of classes in tasks with seen domains versus those with unseen domains\footnote{Unless otherwise specified, we use ``unseen domain'' to denote a domain that has been unseen for a certain task but seen for other tasks. We also examine a ``completely unseen domain'' for all tasks, and present the results in Table~\ref{table.unseen}.} in DGCL (see Fig.~\ref{fig.CIL_DGCL} \textbf{e}-\textbf{h}). Evaluation on seen domains is essentially equivalent to the CIL setting (while using only a small amount of training samples within one domain) as both the training and test sets of each task belong to the same domain, and all baselines show similar trends of comparably strong performance. 
However, the average accuracy on unseen domains deteriorates significantly for most baselines, which is the main cause of their performance degradation in the DGCL setting. This highlights the limited generalization capacity and the difficulty of transferring domain-relevant information across tasks.
We then provide a more detailed analysis regarding different focuses of the evaluated methods, as described below.

\textbf{Representation Learning}: For continual representation learning of $f_\theta (\cdot)$, different PET architectures show mixed performance in CIL, without a clear optimal choice. 
However, in DGCL, using a prompt pool (i.e., L2P and CODA-Prompt) is generally more effective than either task-specific prompts (i.e., DualPrompt, S-Prompt and S-Prompt++) or task-shared prompts (i.e., DualPrompt and LAE). This is because task-specific prompts hinder the transfer of domain-relevant information across tasks, whereas task-shared prompts have limited capacity to accommodate interference over a range of distinct tasks and domains. In contrast, the prompt pool often has a larger capacity of available tuning parameters than either task-specific prompts or task-shared prompts. The prompt ensembling process implicitly disentangles both information and combines them in a weighted manner, therefore facilitating transfer and reducing interference. SLCA benefits from a similar mechanism by possessing adequate capacity (i.e., full parameter tuning) to ensemble representations across tasks and domains effectively.

\textbf{Output Alignment}: Regarding continual output alignment of $h_\psi (\cdot)$, PET-based CL methods with representation recovery generally outperform those without in CIL, as the ensemble of semantic-relevant information rectifies potential biases in predictions. However, this is not necessarily the case in DGCL (e.g., HiDe underperforms L2P in many cases), which underscores the challenges in approximating and recovering domain-relevant information. Notably, methods that incorporate representations with higher complexity tend to perform better, with performance ranking as: ICON (prototypes) $<$ HiDe (multiple centroids) $<$ RanPAC (high-dimensional covariance matrix) $<$ SLCA (multiple covariance matrices). While RanPAC achieves near-perfect accuracy on seen domains, its performance on unseen domains drops drastically. For example, its average in-domain accuracy is 89.25\% whereas average out-domain accuracy is only 1.44\% on Split DigitsDG. This suggests that representation recovery strategies with higher complexity (e.g., second-order statistics) may effectively capture domain-relevant information, but struggle to transfer or even interfere with generalization to other domains.

In summary, state-of-the-art baselines face significant limitations in both representation learning and output alignment. Relatively simple methods instead perform better, since many more advanced designs in conventional CL settings fail to adapt to DGCL.
For representation learning, an ensemble of semantic- and domain-relevant information with adequate capacity in tuning parameters tends to be a desirable choice, such as the prompt pool and full parameter tuning. For output alignment, an ensemble of semantic- and domain-relevant information is also important to obtain balanced and generalized predictions, which requires accurately approximating and recovering feature distributions across tasks and domains. These empirical insights align with how the human brain processes incremental information in diverse forms and generalizes the underlying knowledge to different scenarios (Fig.~\ref{fig.distributed}). We therefore sought a more effective solution to DGCL based on the current progress, which will be presented in the next section.

\section{Method}\label{sec.method}

\begin{figure}[t]
    \centering
    \vspace{-0.1cm}
    \includegraphics[width=0.90\textwidth]{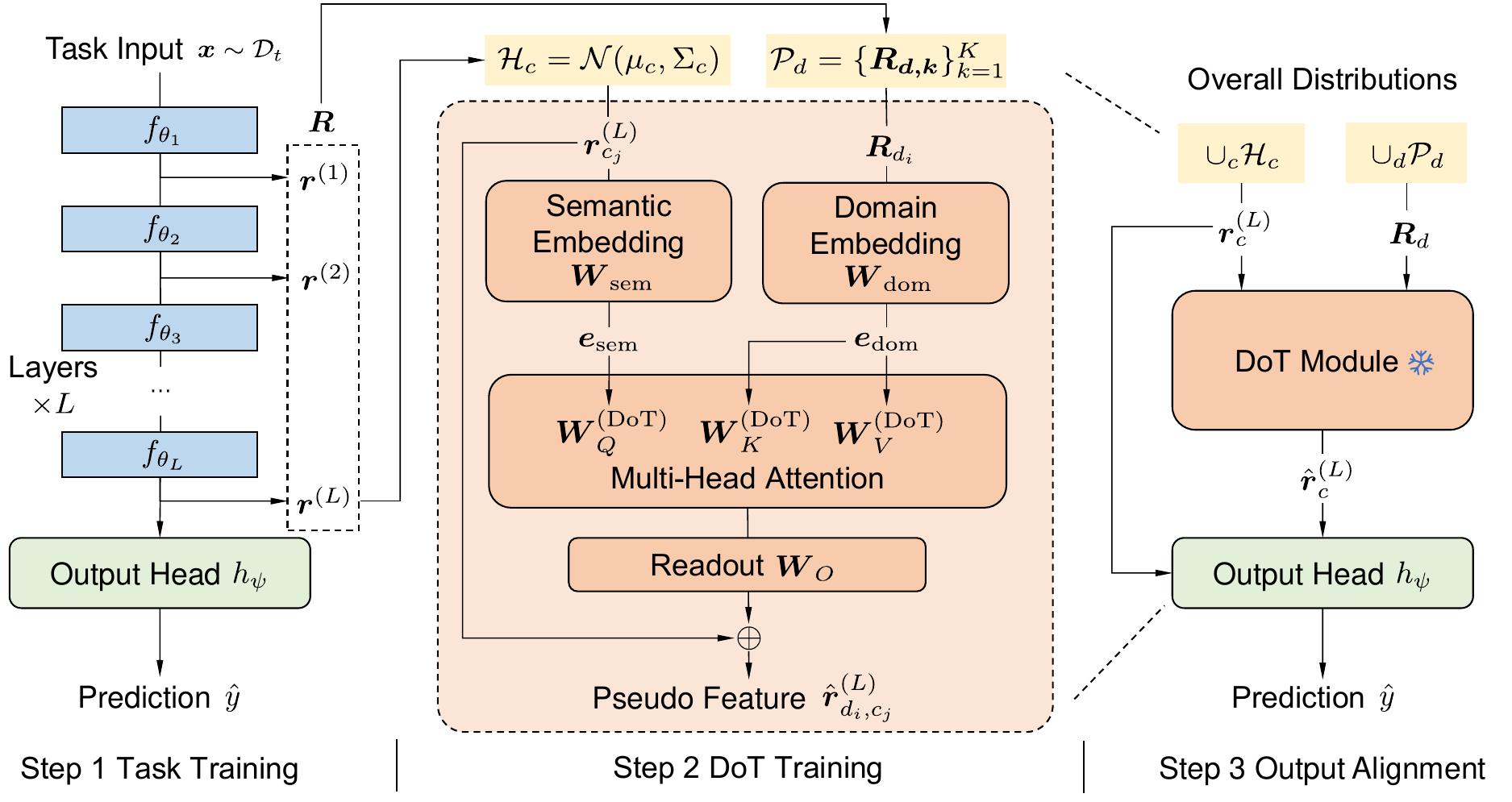}
\caption{Illustration of the proposed framework for DGCL. In step 1, the entire model is trained with incremental tasks, where the feature distributions $\mathcal{H}_c$ and $\mathcal{P}_d$ are preserved. In step 2, the DoT module is trained to generate pseudo feature based on sampled features. In step 3, the output layer $h_\psi$ is aligned with both sampled features and pseudo features produced by the DoT module.}
\vspace{-0.2cm}
\label{fig.dot_method}
\end{figure}

In this section, we draw inspiration from the distributed-plus-hub theory~\cite{patterson2007you} of the human brain, and propose adaptive Domain Transformation (DoT) to address DGCL (see Fig.~\ref{fig.dot_method}). The key innovation is to disentangle semantic- and domain-relevant information in representation learning, and adaptively transform previously learned representations across encountered tasks and domains for output alignment. We first describe an interesting observation that PTMs can naturally disentangle these two types of information in layer-wise features, and then present our approach that strategically exploits this property for domain transformation.

\subsection{Layer-Wise Domain Disentanglement}\label{sec.method_feature}

To formally describe the limitation of existing methods and to propose our solution, we begin with revisiting the basic (continual) learning objective.
Given the training set $\mathcal{D}_t = \{(\boldsymbol{x}_{t,n}, y_{t,n})\}_{n=1}^{N_t}$ of task $t$, the neural network model $h_\psi(f_\theta (\cdot))$ is often optimized with the following loss function: 
\begin{equation}\label{eq.task_loss}
    \mathcal{L}_{t}(\theta, \psi) = \frac{1}{N_t}\sum_{n=1}^{N_t} \mathcal{L}_{\text{task}}(h_\phi(f_\theta(\boldsymbol{x}_{t,n})), y_i),
\end{equation}
where the task-specific loss $\mathcal{L}_{\text{task}}$ can be further specified as the cross-entropy loss $\mathcal{L}_{\text{ce}}$ for classification tasks.
This loss function is to learn the mapping function from $P_X^{(t)} = P_{\text{dom}}^{(t)} \circ P_{\text{sem}}^{(t)}$ to $P_Y^{(t)}$.
Even if the mapping functions of individual tasks are completely preserved from catastrophic forgetting (e.g., by using state-of-the-art CL methods), it is still difficult to address the objective of DGCL, i.e., learning the mapping function from $\bigcup_{i=1}^T P_{\text{dom}}^{(i)} \circ P_{\text{sem}}^{(t)}$ to $P_Y^{(t)}$. Since the domain-relevant factor $P_\text{dom}^{(i)}$ varies in DGCL while the semantic-relevant factor $P_{\text{sem}}^{(t)}$ keeps aligned with $P_Y^{(t)}$, a promising solution is to disentangle $P_\text{dom}^{(i)}$ from $P_{\text{sem}}^{(t)}$, approximate and preserve both distributions, and transform them across encountered tasks and domains. 

However, it is highly nontrivial to explicitly decouple these two factors in DGCL, because the training set for each task and its associated domain is provided sequentially, which is a fundamental difference from multi-source DG. To address this, we propose to implicitly decouple these two factors by exploiting the innate differentiation of layer-wise features in PTMs. Specifically, the pre-trained backbone $f_\theta(\cdot)$ often consists of multiple consecutive layers to encode input data into well-distributed representations. This process can be expressed as $f_\theta(\cdot) = \left(f_{\theta_1}\circ f_{\theta_2}\circ \cdots \circ f_{\theta_L}\right) (\boldsymbol{x})$ of $L$ consecutive layers, generating layer-wise token embeddings $\boldsymbol{h}^{(l)}$. The feature representations $\boldsymbol{r}^{(l)}\in \mathbb{R}^m$ are often extracted as the global average vector or the CLS token of $\boldsymbol{h}^{(l)}$, denoted as $\boldsymbol{r}^{(l)} = g(\boldsymbol{h}^{(l)})$. We further denote all the intermediate features for $1:L$ layers as $\boldsymbol{R}=[\boldsymbol{r}^{(1)}, \boldsymbol{r}^{(2)}, \cdots, \boldsymbol{r}^{(L)}]^{\top} \in \mathbb{R}^{L\times m}$.\footnote{For clarity, the layer identity $l$ of $\boldsymbol{r}$ is placed in superscript. Its underlying task identity $t$, domain identity $d_t$, class identity $c$, and instance identity $n$ are placed in subscript, while omitted if not necessary.}

\begin{figure}[t]
    \centering
    \vspace{-0.1cm}
    \includegraphics[width=1.00\textwidth]{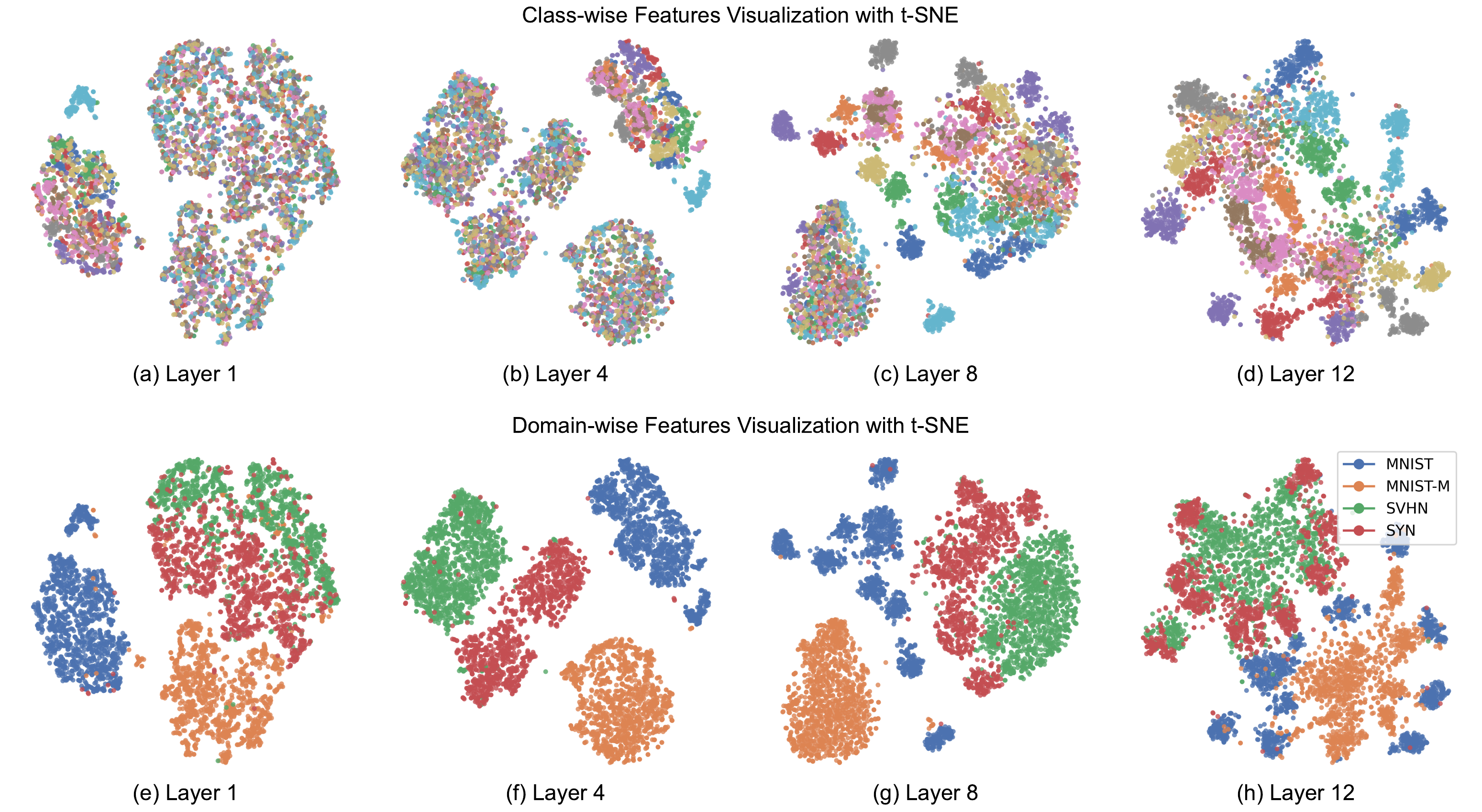}
    \vspace{-0.1cm}
\caption{Layer-wise feature visualization using t-SNE. We extract the layer-wise features $\boldsymbol{r}^{(l)}$ of a ViT-B/16 backbone on the test set of Split DigitsDG, which has been trained using SLCA for all incremental tasks. \textbf{a}-\textbf{d}, Points are colored with their class labels. \textbf{e}-\textbf{h}, Points are colored with their domain identities. Best viewed in color.}
\vspace{-0.2cm}
\label{fig.layer_feature}
\end{figure}

Intuitively, the layer-wise features $\boldsymbol{r}^{(l)}$ closer to $h_\psi(\cdot)$ tend to carry more semantic-relevant information, with the final-layer features $\boldsymbol{r}^{(L)}$ carrying the most. The intermediate features in $\boldsymbol{R}$, on the other hand, carry more domain-relevant information to varying degrees. This trend is validated by our empirical investigation of CL with PTMs under the DGCL setting. As shown in Fig.~\ref{fig.layer_feature}, we perform t-SNE visualization of $\boldsymbol{r}^{(l)}$ for each backbone layer $l \leq L$. 
The intermediate layers (e.g., $l=4,8$) provide domain-discriminative features that can effectively separate different domains, while the deeper layers ($l=12$) allow for semantic discrimination after task-specific fine-tuning. This suggests that the deeper layers capture more semantic-relevant information while the intermediate layers encode more domain-relevant information, aligning with the distributed-plus-hub theory~\cite{patterson2007you} of neural representations (Fig.~\ref{fig.distributed}).
This innate capability of domain disentanglement is then used to capture $P_\text{dom}^{(t)}$ and $P_{\text{sem}}^{(t)}$ for domain transformation, as detailed below.

\subsection{Attention-Based Domain Transformation}\label{sec.method_dot}

With the layer-wise differentiation of semantic- and domain-relevant information, we need to approximate and preserve them for subsequent combination. Considering the distinct properties of their distributions, we adopt adaptive strategies to ensure the effectiveness:

\begin{enumerate}
    \item \textbf{Semantic-Relevant Information}. The final-layer features $\boldsymbol{r}^{(L)}$ collected from $f_\theta(\cdot)$ tend to distribute aligning with the semantic labels. For classification tasks, $\boldsymbol{r}_{c_t}^{(L)}$ belonging to the same class $c_t \in \mathcal{Y}_t$ tend to have a single-peaked distribution~\cite{zhang2023slca, wang2023hierarchical}, which can be naturally approximated as a Gaussian $\mathcal{H}_{c_t}=\mathcal{N}(\boldsymbol{\mu}_{c_t}, \boldsymbol{\Sigma}_{c_t})$ with mean vector $\boldsymbol{\mu}_{c_t}\in \mathbb{R}^m$ and covariance matrix $\boldsymbol{\Sigma}_{c_t}\in\mathbb{R}^{m\times m}$. The covariance matrix can be further simplified into the variance vector by taking the diagonal $\text{Diag}(\boldsymbol{\Sigma}_{c_t})\in\mathbb{R}^{m}$, thus reducing the additional parameter cost to a negligible value.

    \item \textbf{Domain-Relevant Information}. For each task $t$ and its associated domain $d_t$, the intermediate features in $\boldsymbol{R}_{d_t} = [\boldsymbol{r}^{(1)}_{d_t}, \boldsymbol{r}^{(2)}_{d_t}, \cdots, \boldsymbol{r}^{(L)}_{d_t}]^{\top}$ carry domain-relevant information of $d_t$ to varying degrees, which are difficult to fit with a specific probability distribution. Therefore, we simply approximate their distributions $\mathcal{P}_{d_t}$ by selecting $K$ prototypes $\{\boldsymbol{R}_{d_t,k}\}_{k=1}^K$ (through random sampling or $K$-nearest neighbors).
\end{enumerate}

Next, we propose an attention-based strategy to aggregate semantic- and domain-relevant information from these preserved feature distributions.
After sampling the final-layer features $\boldsymbol{r}_{c_j}^{(L)}\sim \mathcal{H}_{c_j}$ of encountered class $c_j\in \mathcal{Y}_j$ and intermediate features $\boldsymbol{R}_{d_i} \sim \mathcal{P}_{d_i}$ of encountered domain $d_i$, we first project them into different latent spaces with learnable semantic embedding matrix $\boldsymbol{W}_\text{sem}\in \mathbb{R}^{m \times m}$ and domain embedding matrix $\boldsymbol{W}_\text{dom}\in \mathbb{R}^{m \times m}$, respectively:
\begin{equation}\label{eq.embed}
    \boldsymbol{e}_\text{sem} = \boldsymbol{r}_{c_j}^{(L)}\boldsymbol{W}_\text{sem} \in \mathbb{R}^m, \quad 
    \boldsymbol{e}_\text{dom} = \boldsymbol{R}_{d_i}\boldsymbol{W}_\text{dom} \in \mathbb{R}^{L\times m}.
\end{equation}
This embedding step further enforces the separation of semantic- and domain-relevant information carried by the original features $\boldsymbol{r}_{c_j}^{(L)}$ and $\boldsymbol{R}_{d_i}$.
Thus, we regard $\boldsymbol{e}_\text{sem}$ as the query vector and $\boldsymbol{e}_\text{dom}$ as both the key and the value vector to perform the attention calculation, so that the extracted information can be flexibly reassembled:
\begin{equation}\label{eq.MSA}
\boldsymbol{a} = \text{Softmax}\left(\boldsymbol{e}_\text{sem} \boldsymbol{W}_{Q}^{\text{(DoT)}}({\boldsymbol{e}_\text{dom} \boldsymbol{W}_{K}^{\text{(DoT)}}})^\top / \sqrt{m}\right) \boldsymbol{e}_\text{dom} \boldsymbol{W}_{V}^{\text{(DoT)}},
\end{equation}
where $\boldsymbol{W}_Q^{\text{(DoT)}} ,\boldsymbol{W}_K^{\text{(DoT)}}$ and $\boldsymbol{W}_V^{\text{(DoT)}}\in \mathbb{R}^{m\times m}$ are learnable projection matrices in the DoT attention module. In practice, we employ a multi-head version by splitting the embedding dimension $m$ into parallel subspaces. Each head independently computes Eq.(\ref{eq.MSA}) within its subspace, and the results are concatenated as the output (the head identity is omitted for clarity).

Finally, the pseudo features $\hat{\boldsymbol{r}}_{d_i,c_j}^{(L)}$ are obtained through a fully-connected readout layer $\boldsymbol{W}_O\in \mathbb{R}^{m \times m}$ with a residual connection and non-linear activation operation $\sigma$:
\begin{equation}\label{eq.read_out}
    \hat{\boldsymbol{r}}_{d_i,c_j}^{(L)} = \sigma\left(\boldsymbol{r}_{c_j}^{(L)} + \boldsymbol{a}\boldsymbol{W}_O\right).
\end{equation}
This transformation step incorporates the domain-relevant information of $d_i$ into the pseudo features $\hat{\boldsymbol{r}}_{d_i,c_j}^{(L)}$, while aligning them with the semantic anchors $\boldsymbol{r}_{c_j}^{(L)}$. Together, we integrate Eq.(\ref{eq.embed}-\ref{eq.read_out}) for the entire DoT process:
\begin{equation}\label{eq.DoT}
    \hat{\boldsymbol{r}}_{d_i,c_j}^{(L)} = \text{DoT}(\boldsymbol{r}_{c_j}^{(L)},\boldsymbol{R}_{d_i}).
\end{equation}

To ensure the generated pseudo features $\hat{\boldsymbol{r}}_{d_i,c_j}^{(L)}$ positioned appropriately in semantic- and domain-specific latent spaces, we introduce two contrastive losses $\mathcal{L}_{\text{cls}}$ and $\mathcal{L}_{\text{dom}}$ to optimize the linear projection heads $p_{\text{cls}}$ and $p_{\text{dom}}$, respectively:
\begin{equation}\label{eq.cls_con_loss}
    \mathcal{L}_{\text{cls}}(\hat{\boldsymbol{r}}_{d_i,c_j}^{(L)}) = - \sum_{\boldsymbol{r}^{(L)}\sim \mathcal{H}_{c_j}}\log \left[\exp\left(p_{\text{cls}}(\hat{\boldsymbol{r}}_{d_i,c_j}^{(L)})\cdot p_{\text{cls}}(\boldsymbol{r}^{(L)})/\tau\right)/s_{\text{cls}}\right],
\end{equation}
\begin{equation}\label{eq.dom_con_loss}
    \mathcal{L}_{\text{dom}}(\hat{\boldsymbol{r}}_{d_i,c_j}^{(L)}) = 
    - \sum_{\boldsymbol{r}_{d_i}^{(L)}\sim \bigcup_{c \in \mathcal{Y}_i}\mathcal{H}_{c}}
    \log \left[\exp\left(p_{\text{dom}}(\hat{\boldsymbol{r}}_{d_i,c_j}^{(L)})\cdot p_{\text{dom}}(\boldsymbol{r}_{d_i}^{(L)})/\tau\right)/s_{\text{dom}}\right],
\end{equation}
where $s_{\text{cls}}$ and $s_{\text{dom}}$ are normalizing factors representing the sum of latent distances between $\hat{\boldsymbol{r}}_{d_i,c_j}^{(L)}$ and all other sampled final-layer features $\boldsymbol{r}^{(L)}\sim \mathcal{H}_{c}$ for $\forall c\in \mathcal{Y}_{1:t}$, and $\tau$ is a temperature hyperparameter. 
$\mathcal{L}_{\text{cls}}$ optimizes the semantic similarity between $\hat{\boldsymbol{r}}_{d_i,c_j}^{(L)}$ and $\boldsymbol{r}^{(L)}$ of the same class $c_j$; 
while $\mathcal{L}_{\text{dom}}$ optimizes the domain similarity between $\hat{\boldsymbol{r}}_{d_i,c_j}^{(L)}$ and $\boldsymbol{r}_{d_i}^{(L)}$ of the same domain $d_i$.

The overall loss function for training the DoT process consists of both $\mathcal{L}_{\text{cls}}$ and $\mathcal{L}_{\text{dom}}$, controlled by a loss-weight hyperparameter $\lambda \in (0, 1)$:
\begin{equation}\label{eq.dot_loss}
    \mathcal{L}_{\text{DoT}}  =  (1-\lambda) \mathcal{L}_{\text{cls}} + \lambda \mathcal{L}_{\text{dom}}.
\end{equation}

By transforming feature representations across all encountered tasks and domains, we perform output alignment with these pseudo features to obtain balanced and generalized predictions.
Specifically, we synthesize $\hat{\boldsymbol{r}}_{d_i, c_j}^{(L)}$ from both $\boldsymbol{r}_{c_j}^{(L)}\sim \mathcal{H}_{c_j}, \forall c_j\in \mathcal{Y}_j$ and $\boldsymbol{R}_{d_i}$ for all possible $(i, j)$ pairs. Therefore, $h_\psi(\cdot)$ is enforced to learn the sampled features $\boldsymbol{r}^{(L)}$ and generated pseudo features $\hat{\boldsymbol{r}}^{(L)}$ with cross-entropy loss: 
\begin{equation}\label{eq.head_loss}
    \mathcal{L}_{\text{OA}} =\sum_{\boldsymbol{r}^{(L)} \sim \bigcup_{c} \mathcal{H}_c} \sum_{\boldsymbol{R} \sim \bigcup_{d} \mathcal{P}_d} \left[\mathcal{L}_{\text{ce}}(h_{\psi}(\boldsymbol{r}^{(L)}), c) + \mathcal{L}_{\text{ce}}(h_{\psi}(\text{DoT}(\boldsymbol{r}^{(L)},\boldsymbol{R})), c)\right].
\end{equation}

After rectifying the output layer with Eq.~\ref{eq.head_loss},
the DoT parameters (i.e., $\boldsymbol{W}_{\text{sem}}$, $\boldsymbol{W}_{\text{dom}}$, $\boldsymbol{W}_{Q}^{\text{(DoT)}}$, $\boldsymbol{W}_{K}^{\text{(DoT)}}$, $\boldsymbol{W}_{V}^{\text{(DoT)}}$, $\boldsymbol{W}_{O}$, $p_{\text{cls}}$ and $p_{\text{dom}}$) can be simply discarded to avoid additional parameter overhead. 
Once a new task is introduced, these parameters are reinitialized and trained with updated $\mathcal{H}_{c_t}$ and $\mathcal{P}_{d_t}$, so as to accommodate the incoming classes and domain. The overall training pipeline is summarized in Alg.~\ref{alg.dot}.

\begin{algorithm}[h]
\caption{Training Algorithm of DoT in DGCL}
\label{alg.dot}
\begin{algorithmic}[1]
\State \textbf{Input}: Task sequence $\{\mathcal{D}_1, ..., \mathcal{D}_T\}$ with domains $\{d_1, ..., d_T\}$
\State \textbf{Hyperparameters}: DoT's training epochs $E_{\text{DoT}}$, OA's training epochs $E_{\text{OA}}$, prototype count $K$, loss weight $\lambda$

\State \textbf{Phase 1: Task Training and Distribution Accumulation}
\State Initialize semantic memory $\mathcal{H} \gets \emptyset$, domain memory $\mathcal{P} \gets \emptyset$
\For{each task $t=1$ to $T$}
    \LineComment{Standard CL training on $\mathcal{D}_t$}
    \State Update $f_\theta, h_\psi$ via $\mathcal{L}_t$ in Eq.(\ref{eq.task_loss})
    
    \LineComment{Accumulate semantic distributions}
    \For{each class $c \in \mathcal{Y}_t$}
        \State $\{\boldsymbol{r}^{(L)}_n\} \gets$ final-layer features of class $c$
        \State $\mathcal{H}_c \gets \mathcal{N}(\text{mean}(\{\boldsymbol{r}^{(L)}_n\}), \text{cov}(\{\boldsymbol{r}^{(L)}_n\}))$
        \State $\mathcal{H} \gets \mathcal{H} \cup \{\mathcal{H}_c\}$
    \EndFor
    
    \LineComment{Accumulate domain prototypes}
    \State $\{\boldsymbol{R}_n\} \gets$ intermediate features from $\mathcal{D}_t$
    \State $\mathcal{P}_{d_t} \gets \text{RandomSample}(\{\boldsymbol{R}_n\}, K)$ 
    \State $\mathcal{P} \gets \mathcal{P} \cup \{\mathcal{P}_{d_t}\}$
\EndFor
\State
\State \textbf{Phase 2: Attention-Based Domain Transformation}
\State Initialize DoT parameters $\boldsymbol{W}_{\text{sem}}$, $\boldsymbol{W}_{\text{dom}}$, $\boldsymbol{W}_Q$, $\boldsymbol{W}_K$, $\boldsymbol{W}_V$, $\boldsymbol{W}_O$
\For{epoch $=1$ to $E_{\text{DoT}}$}
    \State Sample domain $d_i \sim \mathcal{P}$, class $c_j \sim \mathcal{H}$
    \State $\boldsymbol{r}_{c_j}^{(L)} \sim \mathcal{H}_{c_j}$, $\boldsymbol{R}_{d_i} \sim \mathcal{P}_{d_i}$
    
    \LineComment{DoT feature transformation}
    \State $\hat{\boldsymbol{r}} \gets \text{DoT}(\boldsymbol{r}_{c_j}^{(L)}, \boldsymbol{R}_{d_i})$ in Eq.(\ref{eq.embed}-\ref{eq.read_out})
    
    \LineComment{Contrastive alignment}
    \State Compute $\mathcal{L}_{\text{cls}}$ in Eq.(\ref{eq.cls_con_loss}) and $\mathcal{L}_{\text{dom}}$ in Eq.(\ref{eq.dom_con_loss})
    \State Update DoT params via $\nabla \mathcal{L}_{\text{DoT}} = \nabla[(1-\lambda)\mathcal{L}_{\text{cls}} + \lambda\mathcal{L}_{\text{dom}}]$ in Eq.(\ref{eq.dot_loss})
\EndFor
\State
\State \textbf{Phase 3: Output Alignment with Synthesized Features}
\For{epoch $=1$ to $E_{\text{OA}}$}
    \For{each $(d_i, c_j)$ pair in $\mathcal{P} \times \mathcal{H}$}
        \State $\boldsymbol{r}_{c_j}^{(L)} \sim \mathcal{H}_{c_j}$, $\boldsymbol{R}_{d_i} \sim \mathcal{P}_{d_i}$
        \State $\hat{\boldsymbol{r}}^{(L)} \gets \text{DoT}(\boldsymbol{r}_{c_j}^{(L)}, \boldsymbol{R}_{d_i})$
        \State Update $h_\psi$ via $\nabla \mathcal{L}_{\text{OA}}=\nabla[\mathcal{L}_{\text{ce}}(h_\psi(\boldsymbol{r}_{c_j}^{(L)}), c_j)+ \mathcal{L}_{\text{ce}}(h_\psi(\hat{\boldsymbol{r}}^{(L)}), c_j)]$ in Eq.(\ref{eq.head_loss})
    \EndFor
\EndFor

\State \Return Frozen $f_\theta$, aligned $h_\psi$
\end{algorithmic}
\end{algorithm}

\section{Experiment}\label{sec.exp}
In this section, we first describe the experimental setups of DGCL, and then present the experimental results with an extensive analysis.

\subsection{Experimental Setup}\label{sec.exp_setup}

\textbf{Benchmark.} We construct evaluation benchmarks of DGCL with four representative datasets widely used in both DG and CL, including Office-Home~\cite{venkateswara2017deep} of 4 domains (art, clip art, product, and real) with 65 classes, DigitsDG~\cite{zhou2020learning} of 4 domains (MNIST~\cite{lecun1998gradient}, SVHN~\cite{netzer2011reading}, MNIST-M~\cite{ganin2015unsupervised}, and SYN~\cite{ganin2015unsupervised}) with 10 classes; CORe50~\cite{lomonaco2017core50} of 11 domains (changes in lighting, background, and occlusions) with 50 classes, and DomainNet~\cite{peng2019moment} of 6 domains (clip art, infograph, painting, quickdraw, real, and sketch) with 345 classes. The split of training samples and testing samples of each domain and each class follows their original papers. Each dataset is further split into multiple incremental tasks with disjoint classes, where the training samples are restricted to a randomly selected domain while testing samples span all domains. We further ensure that each domain is assigned to the training samples of at least one task, except in the CORe50 dataset where one domain remains unvisited (1 out of 11 domains is excluded for 10 tasks). Detailed statistics of DGCL benchmarks are summarized in Table~\ref{table.dataset_statistics}.

\begin{table*}[th]
\centering
\caption{Detailed statistics of DGCL benchmarks, including the total number of images, domains, and classes, as well as the number of incremental tasks and incremental classes in each task.}
\resizebox{0.90\textwidth}{!}{ 
\begin{tabular}{@{}lccccc@{}}
\toprule
Dataset & Total-Image & Total-Domain & Total-Class & Incre-Task & Incre-Class \\
\midrule
Office-Home~\cite{venkateswara2017deep}    & 15,588    & 4       & 65    & 5     & 13          \\
DigitsDG~\cite{zhou2020learning}       & 24,000    & 4       & 10    & 5     & 2           \\
CORe50~\cite{lomonaco2017core50}         & 160,000   & 11      & 50    & 10    & 5           \\
DomainNet~\cite{peng2019moment}      & 600,000   & 6       & 345   & 10    & 35          \\ \bottomrule
\end{tabular}
}
\label{table.dataset_statistics}
\end{table*}

\textbf{Baseline.} We consider a variety of recent strong baselines for CL with PTMs, especially for PET-based CL methods. A majority of them are designed for CIL, including L2P~\cite{wang2022learning_l2p}, DualPrompt~\cite{wang2022dualprompt}, S-Prompt++~\cite{wang2023hierarchical}, CODA-Prompt~\cite{smith2023coda}, LAE~\cite{gao2023unified}, SLCA~\cite{zhang2023slca}, RanPAC~\cite{mcdonnell2024ranpac}, and HiDe~\cite{wang2023hierarchical,wang2024hide}. 
We also include methods designed for other CL settings, such as S-Prompt~\cite{wang2022sprompts} for DIL and ICON~\cite{park2025versatile} for VIL. The above methods share some convergent ideas, such as task-specific parameters, task-shared parameters, and representation recovery, as summarized in Sec.~\ref{sec.pet_ptms}. While LAE~\cite{gao2023unified} and HiDe~\cite{wang2023hierarchical,wang2024hide} apply to various PET techniques, a majority of these methods focus on ProT and PreT, which serves as our default choice unless otherwise specified.

\textbf{Implementation.} We adopt commonly used implementations of the comparison baselines, with PILOT~\cite{zhou2024continual,sun2023pilot} as the codebase. Specifically, we employ a ViT-B/16 backbone with supervised pre-training of ImageNet-21K. We train all methods with an Adam optimizer ($\beta_1=0.9$, $\beta_2=0.999$) of a cosine-decaying learning rate 0.001, except for SLCA using an SGD optimizer of a head learning rate 0.01 and a backbone learning rate 1e-5. Batch sizes and epoch numbers for each method follow the default setting used in PILOT, and get doubled for the larger dataset DomainNet. We resize the image inputs to $224\times224$ and normalize them to the range $[0,1]$. 
The prompt pools used by L2P and CODA-Prompt follow their default size, prompt length, and location. 
Other PET-based CL methods employ a similar PET architecture, i.e., PreT with prompt length 5 is inserted at layers 1-5. For DoT's configuration, we employ a 4-head attention architecture and assign $K=16$ prototypes per domain distribution $\mathcal{P}_d$. The loss weighting coefficient is set to $\lambda=0.5$. The training epoch number for output alignment is set to $E_{\text{OA}}=3$, mirroring SLCA's implementation. The training epoch number for domain transformation is set to $E_{\text{DoT}}=10$.

\textbf{Evaluation.}
We define the domain-specific accuracy $a_{t,d}$ for task $t$ evaluated on domain $d$, and formulate the following metrics to comprehensively evaluate DGCL performance. 
The average all-domain accuracy $ A_{\text{all}} = \frac{1}{T} \sum_{t=1}^T \frac{1}{|\mathcal{S}|} \sum_{d \in \mathcal{S}} a_{t,d} $ measures the mean accuracy across all tasks and domains, where \(T\) is the total number of tasks and \(\mathcal{S}\) is the set of all encountered domains. 
The average in-domain accuracy $ A_{\text{in}} = \frac{1}{T} \sum_{t=1}^T a_{t,d_t}$ measures the mean accuracy of all tasks, each evaluated on its training domain \(d_t\). 
The average out-domain accuracy $ A_{\text{out}} = \frac{1}{T} \sum_{t=1}^T \frac{1}{|\mathcal{S} \setminus \{d_t\}|} \sum_{d \in \mathcal{S} \setminus \{d_t\}} a_{t,d} $ measures the mean accuracy of all tasks, each evaluated on other domains $d \in \mathcal{S} \setminus \{d_t\}$ out of \(d_t\).
The average worst-case accuracy $ W_{\text{out}} = \frac{1}{T} \sum_{t=1}^T \min_{d \in \mathcal{S} \setminus \{d_t\}} a_{t,d}$ measures the mean accuracy of all tasks, each evaluated on its worst-performing domain out of \(d_t\).
The average unseen-domain accuracy $ A_{\text{un}} = \frac{1}{T} \sum_{t=1}^T a_{t,\bar{d}} $ measures the mean accuracy of all tasks on a completely unseen domain $\bar{d} \notin \mathcal{S}$.
The average all-domain forgetting $ F_{\text{all}} = \frac{1}{T} \sum_{t=1}^T \frac{1}{|\mathcal{S}|} \sum_{d \in \mathcal{S}}(\max_{i \leq T} a_{t,d}^{(i)} - a_{t,d}^{(T)}) $, where $a_{t,d}^{(i)}$ and $a_{t,d}^{(T)}$ denote the accuracy of task $t$ on domain $d$ after learning the $i$-th and the final task $T$, respectively.
The average unseen-domain forgetting $ F_{\text{un}} = \frac{1}{T} \sum_{t=1}^T  (\max_{i \leq T} a_{t,\bar{d}}^{(i)} - a_{t,\bar{d}}^{(T)}) $, where $a_{t,\bar{d}}^{(i)}$ and $a_{t,\bar{d}}^{(T)}$ denote the accuracy of task $t$ on a completely unseen domain $\bar{d}$ after learning the $i$-th and the final task $T$, respectively.

\subsection{Experimental Result}

\newcommand{\tabincell}[2]{\begin{tabular}{@{}#1@{}}#2\end{tabular}}

\begin{table*}[t]
\centering
    \vspace{-0.2cm}
    \caption{Overall performance of DGCL. We report the average all-domain accuracy $A_\text{all}$, average in-domain accuracy $A_\text{in}$, average out-domain accuracy $A_\text{out}$, and average worst-case accuracy $W_\text{out}$. We implement the covariance matrix (Cov) and variance vector (Var) for output alignment in DoT and SLCA.
    All results are averaged over three runs with randomly sampled task sequences. The error bar denotes the standard deviation. 
    } 
      \vspace{-0.1cm}
    \smallskip
      \renewcommand\arraystretch{1.15}
     \addtolength{\tabcolsep}{-2pt}
     
	\resizebox{0.98\textwidth}{!}{ 
      \begin{tabular}{l|cccc|cccc}
	 \hline
        \multirow{2}{*}{\,\,\,\,\,\,\,\, Method} & \multicolumn{4}{c|}{Split Office-Home} & \multicolumn{4}{c}{Split DigitsDG} \\
        & $A_{\text{all}}$ ($\uparrow$) & $A_{\text{in}}$ ($\uparrow$) & $A_{\text{out}}$ ($\uparrow$) & $W_{\text{out}}$ ($\uparrow$) 
        & $A_{\text{all}}$ ($\uparrow$) & $A_{\text{in}}$ ($\uparrow$) & $A_{\text{out}}$ ($\uparrow$) & $W_{\text{out}}$ ($\uparrow$) 
        \\
        \hline
       DualPrompt~\cite{wang2022dualprompt} & $72.74$\tiny{$\pm1.07$} & $85.43$\tiny{$\pm1.90$} & $68.12$\tiny{$\pm0.73$} & $56.04$\tiny{$\pm1.04$}  & $23.07$\tiny{$\pm2.88$} & $29.30$\tiny{$\pm7.51$} & $21.03$\tiny{$\pm1.62$} & $13.14$\tiny{$\pm3.32$}  \\ 
       S-Prompt~\cite{wang2022sprompts} &$61.29$\tiny{$\pm1.17$} & $86.77$\tiny{$\pm2.02$} & $52.60$\tiny{$\pm1.30$} & $33.39$\tiny{$\pm2.72$}  & $11.70$\tiny{$\pm0.94$} & $45.72$\tiny{$\pm3.72$} & $0.36$\tiny{$\pm0.07$} & $0.00$\tiny{$\pm0.00$} \\ 
       S-Prompt++~\cite{wang2023hierarchical} & $72.84$\tiny{$\pm1.03$} & $74.27$\tiny{$\pm1.30$} & $71.86$\tiny{$\pm0.90$} & $55.32$\tiny{$\pm3.57$}  &$13.89$\tiny{$\pm0.84$} & $11.94$\tiny{$\pm4.17$} & $15.16$\tiny{$\pm0.93$} & $3.64$\tiny{$\pm1.80$} \\ 
       CODA-Prompt~\cite{smith2023coda} &$73.31$\tiny{$\pm0.34$} & $81.95$\tiny{$\pm0.87$} & $70.15$\tiny{$\pm0.90$} & $58.42$\tiny{$\pm1.10$}  &$33.49$\tiny{$\pm4.58$} & $53.61$\tiny{$\pm5.76$} & $26.79$\tiny{$\pm4.24$} & $15.31$\tiny{$\pm1.61$}  \\ 
       LAE-PreT~\cite{gao2023unified} & $71.15$\tiny{$\pm1.37$} & $82.88$\tiny{$\pm1.01$} & $67.10$\tiny{$\pm1.75$} & $53.92$\tiny{$\pm2.18$} & $19.62$\tiny{$\pm1.07$} & $23.11$\tiny{$\pm4.18$} & $18.45$\tiny{$\pm2.73$} & $8.78$\tiny{$\pm2.65$}  \\ 
       RanPAC~\cite{mcdonnell2024ranpac} & $77.06$\tiny{$\pm1.52$} & $\textbf{94.11}$\tiny{$\pm1.54$} & $70.91$\tiny{$\pm1.34$} & $56.05$\tiny{$\pm0.68$} & $23.40$\tiny{$\pm1.38$} & $\textbf{89.25}$\tiny{$\pm6.06$} & $1.44$\tiny{$\pm0.70$} & $0.06$\tiny{$\pm0.05$} \\ 
       HiDe-PreT~\cite{wang2023hierarchical} & $74.51$\tiny{$\pm0.82$} & $75.99$\tiny{$\pm0.90$} & $73.48$\tiny{$\pm1.33$} & $56.14$\tiny{$\pm3.80$} & $18.47$\tiny{$\pm1.79$} & $15.42$\tiny{$\pm4.33$} & $22.44$\tiny{$\pm3.63$} & $3.86$\tiny{$\pm3.35$} \\ 
       ICON~\cite{park2025versatile} & $69.78$\tiny{$\pm1.67$} & $69.87$\tiny{$\pm3.59$} & $69.22$\tiny{$\pm2.17$} & $55.48$\tiny{$\pm2.80$} & $12.61$\tiny{$\pm0.93$} & $12.39$\tiny{$\pm1.38$} & $12.69$\tiny{$\pm0.84$} & $0.89$\tiny{$\pm0.19$} \\ 
       \cdashline{1-9}[2pt/2pt]
       L2P~\cite{wang2022learning_l2p} & $74.13$\tiny{$\pm1.35$} & $84.57$\tiny{$\pm3.28$} & $70.36$\tiny{$\pm1.20$} & $57.53$\tiny{$\pm1.01$} & $34.62$\tiny{$\pm4.70$} & $44.14$\tiny{$\pm4.89$} & $31.44$\tiny{$\pm7.54$} & $20.69$\tiny{$\pm9.14$}  \\ 
       \rowcolor{blue!8}
       DoT-L2P (Cov) & $78.30$\tiny{$\pm0.82$} & $90.09$\tiny{$\pm1.79$} & $74.03$\tiny{$\pm0.44$} & $61.18$\tiny{$\pm0.68$} & $38.17$\tiny{$\pm1.04$} & $81.97$\tiny{$\pm3.35$} & $23.57$\tiny{$\pm2.11$} & $10.17$\tiny{$\pm3.48$} \\ 
       \rowcolor{green!10}
       DoT-L2P (Var) & $77.63$\tiny{$\pm0.90$} & $88.60$\tiny{$\pm1.71$} & $73.81$\tiny{$\pm0.62$} & $61.09$\tiny{$\pm0.74$} & $38.01$\tiny{$\pm1.11$} & $80.53$\tiny{$\pm3.32$} & $23.84$\tiny{$\pm0.94$} & $9.67$\tiny{$\pm3.22$} \\ 
       \cdashline{1-9}[2pt/2pt]
       SLCA (Cov)~\cite{zhang2023slca} & $78.46$\tiny{$\pm0.31$} & $91.30$\tiny{$\pm0.90$} & $73.60$\tiny{$\pm0.45$} & $62.14$\tiny{$\pm2.15$} & $51.81$\tiny{$\pm2.14$} & $83.11$\tiny{$\pm6.78$} & $41.38$\tiny{$\pm0.85$} & $21.83$\tiny{$\pm2.44$} \\ 
       SLCA (Var)~\cite{zhang2023slca} & $77.89$\tiny{$\pm 0.47$} & $89.87$\tiny{$\pm 1.28$} & $73.34$\tiny{$\pm 0.99$} & $62.16$\tiny{$\pm 1.87$} & $56.39$\tiny{$\pm 4.56$} & $82.78$\tiny{$\pm 5.52$} & $47.59$\tiny{$\pm 4.50$} & $28.94$\tiny{$\pm 6.54$}  \\ 
       \rowcolor{blue!8}
       DoT-SLCA (Cov) & $\textbf{79.77}$\tiny{$\pm1.01$} & $91.89$\tiny{$\pm1.56$} & $\textbf{75.23}$\tiny{$\pm0.91$} & $63.83$\tiny{$\pm0.66$} & $61.68$\tiny{$\pm2.83$} & $87.92$\tiny{$\pm2.10$} & $52.93$\tiny{$\pm3.13$} & $32.39$\tiny{$\pm5.76$} \\ 
       \rowcolor{green!10}
       DoT-SLCA (Var) & $79.41$\tiny{$\pm 0.44$} & $90.66$\tiny{$\pm 1.66$} & $75.17$\tiny{$\pm 0.15$} & $\textbf{64.73}$\tiny{$\pm 0.81$} & $\textbf{62.48}$\tiny{$\pm 3.20$} & $87.28$\tiny{$\pm 1.44$} & $\textbf{54.21}$\tiny{$\pm 4.58$} & $\textbf{35.14}$\tiny{$\pm 9.34$} \\ 
       \hline
        \hline
        \multirow{2}{*}{\,\,\,\,\,\,\,\, Method} & \multicolumn{4}{c|}{Split CORe50} & \multicolumn{4}{c}{Split DomainNet} \\
        & $A_{\text{all}}$ ($\uparrow$) & $A_{\text{in}}$ ($\uparrow$) & $A_{\text{out}}$ ($\uparrow$) & $W_{\text{out}}$ ($\uparrow$) 
        & $A_{\text{all}}$ ($\uparrow$) & $A_{\text{in}}$ ($\uparrow$) & $A_{\text{out}}$ ($\uparrow$) & $W_{\text{out}}$ ($\uparrow$) 
        \\
        \hline
       DualPrompt~\cite{wang2022dualprompt} & $52.76$\tiny{$\pm2.27$} & $81.55$\tiny{$\pm3.75$} & $49.82$\tiny{$\pm2.23$} & $35.49$\tiny{$\pm1.83$}   & $32.01$\tiny{$\pm1.88$} & $48.65$\tiny{$\pm4.35$} & $24.38$\tiny{$\pm2.75$} & $4.42$\tiny{$\pm0.32$} \\ 
       S-Prompt~\cite{wang2022sprompts} & $28.75$\tiny{$\pm4.51$} & $69.93$\tiny{$\pm10.36$} & $24.48$\tiny{$\pm4.07$} & $5.06$\tiny{$\pm2.75$}   & $14.53$\tiny{$\pm0.65$} & $47.30$\tiny{$\pm2.84$} & $6.02$\tiny{$\pm1.53$} & $0.04$\tiny{$\pm0.03$} \\ 
       S-Prompt++~\cite{wang2023hierarchical} &$35.66$\tiny{$\pm1.89$} & $34.31$\tiny{$\pm2.03$} & $35.59$\tiny{$\pm1.16$} & $16.77$\tiny{$\pm1.94$}  &$29.92$\tiny{$\pm0.68$} & $27.71$\tiny{$\pm0.64$} & $29.57$\tiny{$\pm1.12$} & $\textbf{13.50}$\tiny{$\pm0.96$} \\ 
       CODA-Prompt~\cite{smith2023coda}& $56.83$\tiny{$\pm2.33$} & $84.64$\tiny{$\pm3.53$} & $53.80$\tiny{$\pm2.37$} & $35.96$\tiny{$\pm3.23$}     &$36.07$\tiny{$\pm1.97$} & $57.23$\tiny{$\pm1.83$} & $26.94$\tiny{$\pm3.22$} & $4.66$\tiny{$\pm0.16$} \\ 
       LAE-PreT~\cite{gao2023unified} & $55.47$\tiny{$\pm1.87$} & $80.09$\tiny{$\pm2.66$} & $52.71$\tiny{$\pm2.00$} & $39.03$\tiny{$\pm0.41$} & $34.18$\tiny{$\pm0.28$} & $54.53$\tiny{$\pm1.27$} & $26.17$\tiny{$\pm2.09$} & $4.71$\tiny{$\pm0.84$}  \\ 
       RanPAC~\cite{mcdonnell2024ranpac} & $56.90$\tiny{$\pm0.16$} & $\textbf{100.00}$\tiny{$\pm0.00$} & $52.34$\tiny{$\pm0.32$} & $18.27$\tiny{$\pm2.77$} & $39.96$\tiny{$\pm2.19$} & $\textbf{67.77}$\tiny{$\pm0.56$} & $30.57$\tiny{$\pm4.61$} & $2.15$\tiny{$\pm1.39$} \\ 
       HiDe-PreT~\cite{wang2023hierarchical} &$32.87$\tiny{$\pm3.17$} & $30.96$\tiny{$\pm4.86$} & $32.78$\tiny{$\pm3.06$} & $14.75$\tiny{$\pm2.56$}  & $31.80$\tiny{$\pm3.41$} & $24.47$\tiny{$\pm4.55$} & $26.46$\tiny{$\pm4.67$} & $12.55$\tiny{$\pm1.59$} \\ 
       ICON~\cite{park2025versatile} & $27.07$\tiny{$\pm4.99$} & $22.19$\tiny{$\pm6.77$} & $27.56$\tiny{$\pm4.81$} & $7.93$\tiny{$\pm5.37$} & $30.09$\tiny{$\pm0.90$} & $28.61$\tiny{$\pm0.56$} & $30.27$\tiny{$\pm1.43$} & $13.36$\tiny{$\pm0.19$}  \\ 
       \cdashline{1-9}[2pt/2pt]
       L2P~\cite{wang2022learning_l2p} & $55.41$\tiny{$\pm3.45$} & $78.20$\tiny{$\pm4.33$} & $53.00$\tiny{$\pm3.20$} & $43.17$\tiny{$\pm2.29$}    & $33.77$\tiny{$\pm1.42$} & $50.04$\tiny{$\pm3.55$} & $26.21$\tiny{$\pm1.48$} & $4.36$\tiny{$\pm1.00$} \\ 
       \rowcolor{blue!8}
       DoT-L2P (Cov) & $73.50$\tiny{$\pm0.41$} & $98.33$\tiny{$\pm0.68$} & $70.87$\tiny{$\pm0.55$} & $57.20$\tiny{$\pm1.66$} & $38.65$\tiny{$\pm0.77$} & $56.86$\tiny{$\pm2.24$} & $30.56$\tiny{$\pm0.60$} & $4.62$\tiny{$\pm1.03$} \\ 
       \rowcolor{green!10}
       DoT-L2P (Var) & $71.63$\tiny{$\pm0.89$} & $96.68$\tiny{$\pm0.68$} & $68.98$\tiny{$\pm1.13$} & $56.11$\tiny{$\pm0.92$} & $38.63 $\tiny{$\pm 0.80$} & $ $56.50\tiny{$\pm 2.20$} & $ 30.62$\tiny{$\pm 0.52$} & $4.68 $\tiny{$\pm 1.01$}    \\ 
       \cdashline{1-9}[2pt/2pt]
       SLCA (Cov)~\cite{zhang2023slca} & $82.29$\tiny{$\pm2.12$} & $98.94$\tiny{$\pm0.23$} & $80.48$\tiny{$\pm2.36$} & $68.09$\tiny{$\pm3.77$} & $43.25$\tiny{$\pm0.62$} & $66.79$\tiny{$\pm1.00$} & $33.91$\tiny{$\pm1.52$} & $4.74$\tiny{$\pm0.51$}  \\ 
       SLCA (Var)~\cite{zhang2023slca} & $79.40$\tiny{$\pm 1.36$} & $97.46$\tiny{$\pm 0.15$} & $77.45$\tiny{$\pm 1.56$} & $65.88$\tiny{$\pm 0.84$} & $43.19$\tiny{$\pm 0.79$} & $65.96$\tiny{$\pm 0.89$} & $33.36$\tiny{$\pm 1.32$} & $5.69$\tiny{$\pm 0.44$}  \\ 
       \rowcolor{blue!8}
       DoT-SLCA (Cov) & $\textbf{84.73}$\tiny{$\pm1.29$} & $99.20$\tiny{$\pm0.22$} & $\textbf{83.14}$\tiny{$\pm1.48$} & $\textbf{73.00}$\tiny{$\pm2.30$}  & $43.82$\tiny{$\pm1.00$} & $66.83$\tiny{$\pm0.89$} & $34.52$\tiny{$\pm1.41$} & $5.10$\tiny{$\pm0.63$} \\ 
       \rowcolor{green!10}
       DoT-SLCA (Var) & $82.05$\tiny{$\pm 2.24$} & $98.21$\tiny{$\pm 0.88$} & $80.28$\tiny{$\pm 2.53$} & $69.99$\tiny{$\pm 3.69$} & $\textbf{44.46}$\tiny{$\pm 0.95$} & $65.96$\tiny{$\pm 1.05$} & $\textbf{35.64}$\tiny{$\pm 1.14$} & $5.88$\tiny{$\pm 0.40$}  \\ 
       \hline
	\end{tabular}
	} 
	\label{table.overall}
	\vspace{-0.2cm}
\end{table*}

\newcommand{\gray}[1]{\color{gray}#1}
\begin{table*}[t]
\centering
    \vspace{-0.2cm}
    \caption{Comparison of computational and parameter overhead. We present the results on Split DigitsDG ($\mathcal{H}_c$ and $\mathcal{P}_d$ for other benchmarks will vary with the number of classes and domains, but the trend is consistent). The training time is evaluated with one-card 3090 GPU, AMD EPYC 7402 (2.8G Hz). $^\dagger$$\mathcal{H}_c$ and $\mathcal{P}_d$ are preserved but \textbf{not} trainable in DGCL. $^\ddagger$The DoT parameters are trainable but \textbf{not} preserved in DGCL.} 
      \vspace{-0.1cm}
    \smallskip
      \renewcommand\arraystretch{1.15}
     \addtolength{\tabcolsep}{-2pt}
\resizebox{0.75\textwidth}{!}{ 
\begin{tabular}{lcccccc}
\toprule
\,\,\,\,\,\,\, Method & $A_{\text{all}}(\%)$ & $\mathcal{H}_c$$^\dagger$ & $\mathcal{P}_d$$^\dagger$ & DoT$^\ddagger$ & Trainable & Training Time \\
\midrule
L2P~\cite{wang2022learning_l2p} & 34.62 & -- & -- & -- & 0.14M & 9.23min \\
\rowcolor{blue!8}
DoT-L2P (Cov) & 38.17 & 5.63M & 1.41M & \gray{2.39M} & 2.53M & 18.25min\\
\rowcolor{green!10}
DoT-L2P (Var) & 38.01 & 0.01M & 1.41M & \gray{2.39M} & 2.53M & 18.20min \\
\cdashline{1-7}[2pt/2pt]
SLCA (Cov)~\cite{zhang2023slca} & 51.81 & 5.63M & -- & -- & 81.82M & 40.41min \\
SLCA (Var)~\cite{zhang2023slca} & 56.39 & 0.01M & -- & -- & 81.82M & 39.88min \\
\rowcolor{blue!8}
DoT-SLCA (Cov) & 61.68 & 5.63M & 1.41M & \gray{2.39M} & 84.21M & 42.41min \\
\rowcolor{green!10}
DoT-SLCA (Var) & 62.48 & 0.01M & 1.41M & \gray{2.39M} & 84.21M & 41.37min \\
\bottomrule
\end{tabular}
} 
	\label{table.efficiency}
	\vspace{-0.1cm}
\end{table*}

\textbf{Overall Performance}: We first present the overall performance of DGCL in Table~\ref{table.overall}. Consistent with the analysis in Sec.~\ref{sec.dgcl_exp}, advanced PET-based CL methods exhibit varying degrees of performance degradation in $A_{\text{all}}$\footnote{We consider $A_{\text{all}}$ as the primary evaluation metric, as it accommodates the overall objective of DGCL in both seen domains and unseen domains.}, especially for unseen domains in $A_{\text{out}}$ compared to seen domains in $A_{\text{in}}$. L2P and CODA-Prompt generally achieve the highest overall performance in this avenue, since they both employ the ensemble of a prompt pool. SLCA significantly outperforms all PET-based CL methods due to the large capacity of full parameter tuning and the representation recovery for output alignment. In comparison, our proposed DoT serves as a plug-in strategy that significantly improves the performance of state-of-the-art CL baselines under both PET (i.e., L2P) and full parameter tuning (i.e., SLCA) paradigms. Notably,
DoT-SLCA improves SLCA mainly in $A_{\text{out}}$ rather than $A_{\text{in}}$, suggesting that the benefits of domain transformation are specific to DGCL rather than CIL. 
DoT-L2P improves L2P in both $A_{\text{in}}$ and $A_{\text{out}}$ due to the additional benefits of output alignment to representation learning, as analyzed in previous work~\cite{zhang2023slca,zhang2024slca++}. 

Since SLCA approximates the distribution of each class as a Gaussian with a dedicated mean vector and covariance matrix, we use it as the default implementation of DoT-SLCA as well as DoT-L2P to ensure comparison clarity and fairness (denoted as ``Cov'' in Table~\ref{table.overall}). However, this implementation incurs parameter overhead of a remarkably high complexity $O(m^2)$~\cite{wang2024hide}, where $m$ denotes the embedding dimension. For example, Office-Home has a total of 65 classes, and the additional parameter overhead is 36.56M with $m=768$ (around 42.51\% of the ViT-B/16 backbone). In this regard, we reduce the covariance matrix to a lightweight variance vector of complexity $O(m)$ (denoted as ``Var'' in Table~\ref{table.overall}), which achieves comparable or even better performance with our proposed DoT in DGCL. This implementation results in additional parameter overhead of only 0.05M for Office-Home (around 0.06\% of the ViT-B/16 backbone), which is essentially negligible. Accordingly, DoT-SLCA (Var) achieves state-of-the-art performance in DGCL and requires much fewer parameters than the original SLCA (Cov) (see Table~\ref{table.efficiency}).

\begin{figure}[t]
    \centering
    \vspace{-0.1cm}
    \includegraphics[width=0.95\textwidth]{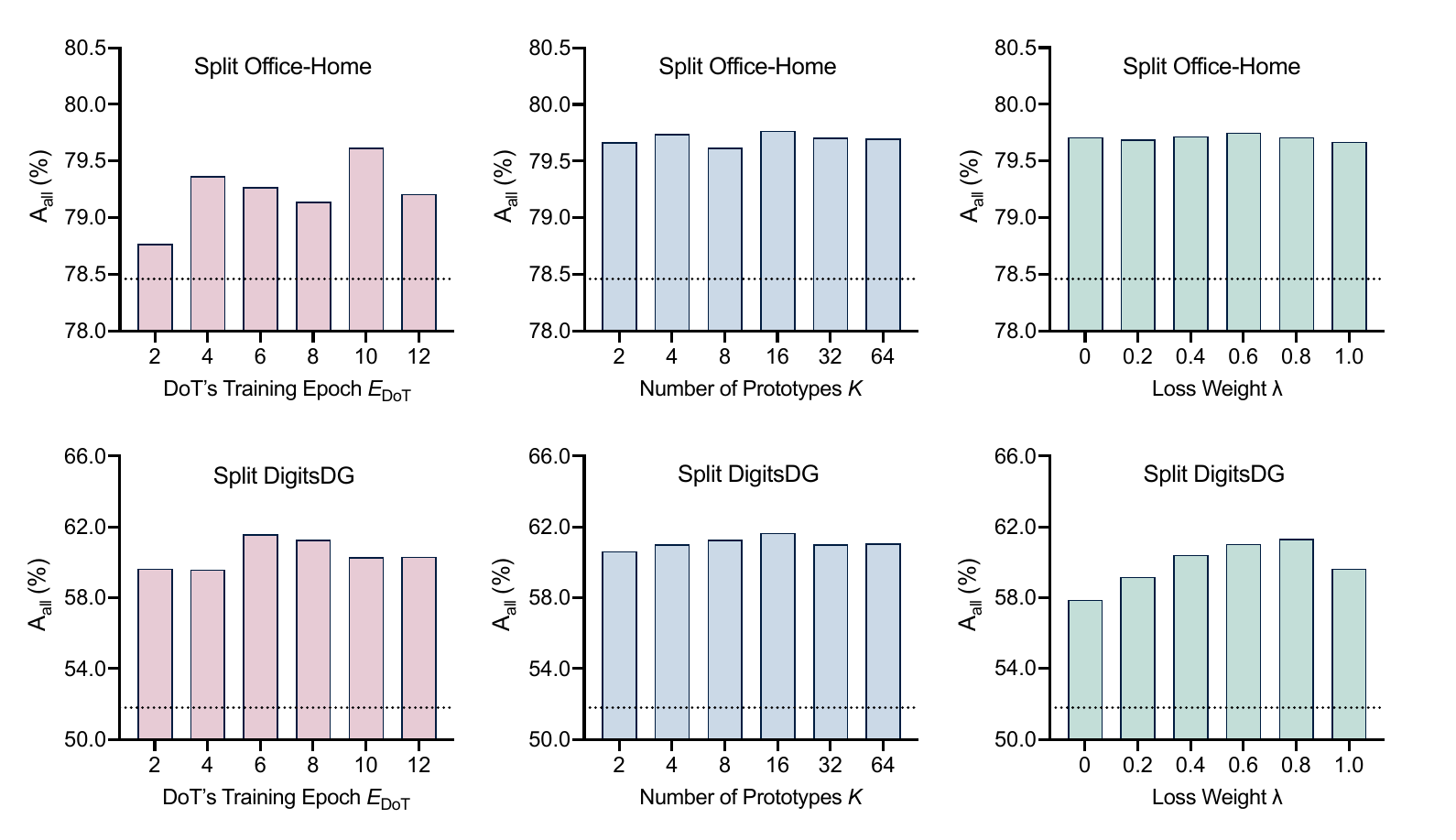}
\caption{Impact of hyperparameters. We evaluate the impact of DoT's training epochs $E_{\text{DoT}}$, number of preserved prototypes $K$, and loss weight $\lambda$ with DoT-SLCA (Cov). The dashed line is SLCA (Cov) as the baseline. All results are averaged over three runs with randomly sampled task sequences. }
\vspace{-0.2cm}
\label{fig.hyperparameters}
\end{figure}

We provide a more detailed analysis of computational and parameter overhead in Table~\ref{table.efficiency}. 
Notably, when DoT performs DGCL, the feature distributions are preserved but not trainable, and the DoT parameters are trainable but not preserved. The simplified variance vector further reduces the cost.
Using an SLCA-like output alignment incurs significant computational overhead (DoT-L2P versus L2P), while the training of DoT parameters is clearly lightweight (DoT-SLCA versus SLCA). As a result, our approach requires only a small amount of additional overhead while achieving strong improvements in DGCL. The additional overhead can be even further reduced by using fewer epochs and prototypes, as shown later in the analysis of Fig.~\ref{fig.hyperparameters}.

\begin{table*}[t]
\centering
    \vspace{-0.2cm}
    \caption{Performance of additional unseen domain and forgetting measurement. We report the average all-domain accuracy $A_\text{all}$ (removing the completely unseen domain), average unseen-domain accuracy $A_\text{un}$, average all-domain forgetting $F_\text{all}$ (removing the completely unseen domain), and average unseen-domain forgetting $F_\text{un}$. We implement the covariance matrix (Cov) and variance vector (Var) for output alignment in DoT and SLCA. All results are averaged over three runs with randomly sampled task sequences. The error bar denotes the standard deviation.
    } 
      \vspace{-0.1cm}
    \smallskip
      \renewcommand\arraystretch{1.15}
     \addtolength{\tabcolsep}{-2pt}
     
	\resizebox{0.98\textwidth}{!}{ 
      \begin{tabular}{l|cccc|cccc}
	 \hline
        \multirow{2}{*}{\,\,\,\,\,\,\,\, Method} & \multicolumn{4}{c|}{Split CORe50} & \multicolumn{4}{c}{Split DigitsDG-C} \\
        & $A_{\text{all}}$ ($\uparrow$) & $A_{\text{un}}$ ($\uparrow$) & $F_{\text{in}}$ ($\downarrow$) & $F_{\text{out}}$ ($\downarrow$) 
        & $A_{\text{all}}$ ($\uparrow$) & $A_{\text{un}}$ ($\uparrow$) & $F_{\text{in}}$ ($\downarrow$) & $F_{\text{out}}$ ($\downarrow$) 
        \\
        \hline
       DualPrompt~\cite{wang2022dualprompt} & $52.99$\tiny{$\pm2.06$} & $49.86$\tiny{$\pm5.25$} & $17.34$\tiny{$\pm1.89$} & $18.36$\tiny{$\pm4.53$}  & $26.08$\tiny{$\pm 2.45$} & $17.69$\tiny{$\pm 1.93$} & $20.85 $\tiny{$\pm 8.84$} & $16.83$\tiny{$\pm 11.76$}  \\ 
       S-Prompt~\cite{wang2022sprompts} & $27.55$\tiny{$\pm 4.35$} & $39.20$\tiny{$\pm 8.64$} & $11.76$\tiny{$\pm 5.88$} & $7.34$\tiny{$\pm 3.46$}  & $14.71$\tiny{$\pm 1.77$} & $12.64$\tiny{$\pm 1.41$} & $10.85$\tiny{$\pm 0.61$} & $15.83$\tiny{$\pm 4.47$} \\ 
       S-Prompt++~\cite{wang2023hierarchical} & $34.79$\tiny{$\pm 1.27$} & $ 42.22 $\tiny{$\pm 1.65$} & $ 9.89 $\tiny{$\pm 0.40$} & $ 9.91 $\tiny{$\pm 0.99$}  & $15.99$\tiny{$\pm 1.77$} & $12.17$\tiny{$\pm 2.40$} & $14.66$\tiny{$\pm 2.84$} & $18.36$\tiny{$\pm 8.65$} \\ 
       CODA-Prompt~\cite{smith2023coda} & $57.41$\tiny{$\pm 2.46$} & $48.58$\tiny{$\pm 2.44$} & $6.45$\tiny{$\pm 1.48$} & $6.17$\tiny{$\pm 1.62$}  & $28.54$\tiny{$\pm 3.32$} & $25.50$\tiny{$\pm 7.09$} & $15.04$\tiny{$\pm 3.86$} & $13.72$\tiny{$\pm 2.65$} \\ 
       LAE-PreT~\cite{gao2023unified} & $55.48$\tiny{$\pm 1.83$} & $52.38$\tiny{$\pm 4.84$} & $17.51$\tiny{$\pm 3.00$} & $19.17$\tiny{$\pm 4.11$}  & $21.89$\tiny{$\pm 1.01$} & $20.92$\tiny{$\pm 3.26$} & $20.88$\tiny{$\pm 5.85$} & $17.83$\tiny{$\pm 9.21$} \\ 
       RanPAC~\cite{mcdonnell2024ranpac} & $55.38$\tiny{$\pm 0.19$} & $69.72$\tiny{$\pm 2.23$} & $18.16$\tiny{$\pm 2.48$} & $11.25$\tiny{$\pm 4.26$}  & $28.54$\tiny{$\pm 3.32$} & $25.50$\tiny{$\pm 7.09$} & $15.04$\tiny{$\pm 3.86$} & $13.72$\tiny{$\pm 2.65$} \\ 
       HiDe-PreT~\cite{wang2023hierarchical} & $32.21$\tiny{$\pm 3.10$} & $36.70$\tiny{$\pm 4.56$} & $\textbf{4.03}$\tiny{$\pm 0.61$} & $\textbf{3.75}$\tiny{$\pm 0.54$}  & $26.45$\tiny{$\pm 2.24$} & $17.97$\tiny{$\pm 4.59$} & $\textbf{2.37}$\tiny{$\pm 1.73$} & $\textbf{2.22}$\tiny{$\pm 0.85$} \\ 
       ICON~\cite{park2025versatile} & $26.24$\tiny{$\pm 5.17$} & $35.37$\tiny{$\pm 4.64$} & $13.37$\tiny{$\pm 1.06$} & $10.92$\tiny{$\pm 3.88$} & $12.31$\tiny{$\pm 2.61$} & $10.83$\tiny{$\pm 0.51$} & $9.71$\tiny{$\pm 7.13$} & $8.86$\tiny{$\pm 1.97$} \\ 
       \cdashline{1-9}[2pt/2pt]
       L2P~\cite{wang2022learning_l2p} & $55.41$\tiny{$\pm 3.63$} & $53.43$\tiny{$\pm 0.49$} & $15.27$\tiny{$\pm 2.10$} & $15.11$\tiny{$\pm 3.23$}  & $32.80$\tiny{$\pm 2.98$} & $27.69$\tiny{$\pm 2.17$} & $19.52$\tiny{$\pm 0.32$} & $16.83$\tiny{$\pm 6.69$} \\ 
       \rowcolor{blue!8}
       DoT-L2P (Cov)  &$73.43$\tiny{$\pm0.16$} & $72.73$\tiny{$\pm3.76$} & $9.06$\tiny{$\pm0.48$} & $8.80$\tiny{$\pm0.67$} & $38.29$\tiny{$\pm 2.21$} & $31.67$\tiny{$\pm 5.51$} & $20.14$\tiny{$\pm 2.01$} & $21.19$\tiny{$\pm 3.52$}\\ 
       \rowcolor{green!10}
       DoT-L2P (Var)  & $71.38$\tiny{$\pm 0.76$} & $69.03$\tiny{$\pm 2.05$} & $10.89$\tiny{$\pm 1.05$} & $11.23$\tiny{$\pm 1.72$} & $38.06$\tiny{$\pm 2.53$} & $29.75$\tiny{$\pm 5.51$} & $19.30$\tiny{$\pm 3.21$} & $22.28$\tiny{$\pm 4.22$}\\ 
       \cdashline{1-9}[2pt/2pt]
       SLCA (Cov)~\cite{zhang2023slca} & $82.23$\tiny{$\pm 2.28$} & $80.48$\tiny{$\pm 3.18$} & $7.27$\tiny{$\pm 1.89$} & $7.46$\tiny{$\pm 1.69$} & $47.62$\tiny{$\pm 3.17$} & $45.11$\tiny{$\pm 13.50$} & $20.42$\tiny{$\pm 5.35$} & $16.11$\tiny{$\pm 7.40$} \\ 
       SLCA (Var)~\cite{zhang2023slca} & $80.55$\tiny{$\pm 0.96$} & $78.11$\tiny{$\pm 2.75$} & $6.60$\tiny{$\pm 0.41$} & $6.82$\tiny{$\pm 2.63$} & $48.57$\tiny{$\pm 6.05$} & $ 42.28 $\tiny{$\pm 14.31$} & $ 15.31 $\tiny{$\pm 0.40$} & $ 10.03 $\tiny{$\pm 6.90$} \\ 
       \rowcolor{blue!8}
       DoT-SLCA (Cov) & $\textbf{84.74}$\tiny{$\pm 1.26$} & $\textbf{83.14}$\tiny{$\pm 2.36$} & $5.99$\tiny{$\pm 0.47$} & $5.67$\tiny{$\pm 0.65$}  & $58.74$\tiny{$\pm 4.95$} & $54.75$\tiny{$\pm 5.62$} & $20.43$\tiny{$\pm 1.94$} & $17.45$\tiny{$\pm 5.86$} \\ 
       \rowcolor{green!10}
       DoT-SLCA (Var)  & $82.14 $\tiny{$\pm 2.16 $} & $ 79.58 $\tiny{$\pm 4.01 $} & $ 6.07 $\tiny{$\pm 0.49 $} & $6.22$\tiny{$\pm 1.79$} & $\textbf{61.72}$\tiny{$\pm 6.35$} & $\textbf{56.69}$\tiny{$\pm 12.02$} & $ 14.88 $\tiny{$\pm 3.07$} & $13.81$\tiny{$\pm 4.96$}\\ 
       \hline
	\end{tabular}
	} 
	\label{table.unseen}
	\vspace{-0.2cm}
\end{table*}

\textbf{Impact of Unseen Domain.}
The DGCL setting emphasizes that the model learns each task from a single domain, while generalizing to all encountered domains out of the task-specific training set. This consideration stems from the fact that there are usually many more tasks (or classes) to be learned than their available domains, consistent with the statistics of widely-used datasets in both DG and CL (Table~\ref{table.dataset_statistics}). We further evaluate the capability of DG under more stringent conditions, i.e., the model should generalize all previously learned tasks to a completely unseen domain after DGCL. For example, for Split CORe50 that consists of 10 tasks and 11 domains, we learn each task with a distinct domain and additionally evaluate the average accuracy of these 10 tasks under the 11th domain. For Split DigitsDG that consists of 5 tasks and 4 domains, we reconstruct this benchmark as learning the 5 tasks with each belonging to one of the 3 domains (i.e., there are several tasks learned from the same domain), denoted as Split DigitsDG-C (``-C'' stands for ``crowded''), and additionally evaluate the average accuracy of these 5 tasks under the 4th domain.

In this way, Split CORe50 and Split DigitsDG-C represent two extreme cases of DGCL: domains being overly sparse (i.e., different domains for each task) versus domains being overly dense (i.e., identical domains for many tasks), allowing for a more comprehensive analysis. As shown in Table~\ref{table.unseen}, our approach provides consistently strong improvements to L2P and SLCA in both cases. Interestingly, the improvement to the average accuracy of unseen domains (i.e., $A_{\text{un}}$) tends to be even more significant than that of all observed domains (i.e., $A_{\text{all}}$), as our cross-task and cross-domain transformation enables PTMs to accumulate domain generalizable knowledge from DGCL. 
We further evaluate the average all-domain forgetting $F_{\text{all}}$ for the maximum decrease of all-domain accuracy and average unseen-domain forgetting $F_{\text{un}}$ for the maximum decrease of unseen-domain accuracy in DGCL. In general, our approach achieves comparable $F_{\text{all}}$ and $F_{\text{un}}$ compared to the corresponding baselines, which further suggests that the improvement to the overall performance is mainly from better generalization to unseen domains. Again, the covariance and variance versions of our approach exhibit comparable performance.

\textbf{Detailed Analysis of DoT.} 
To provide a more in-depth analysis of our approach, we investigate the impact of three critical hyperparameters, including DoT's training epochs $E_{\text{DoT}}$, the number of preserved prototypes $K$, and the loss weight $\lambda$. As shown in Fig.~\ref{fig.hyperparameters}, these hyperparameters are relatively stable across a wide range of specific values, and deliver consistent improvements over the baseline, i.e., DoT-SLCA (Cov) versus SLCA (Cov). This property ensures the robustness of DoT in response to uncertain changes in real-world applications. The results for all hyperparameters exhibit an overall single-peak pattern, especially in Split DigitsDG that enjoys comparably substantial improvements, suggesting the effectiveness of our designs. In particular, we observe that DoT can already achieve strong improvements by using only 4 epochs and preserving only 2 prototypes (our default setting is 10 epochs and 16 prototypes), which further reduces the computational and parameter overhead.

\begin{figure}[t]
    \centering
    \vspace{-0.1cm}
    \includegraphics[width=1.00\textwidth]{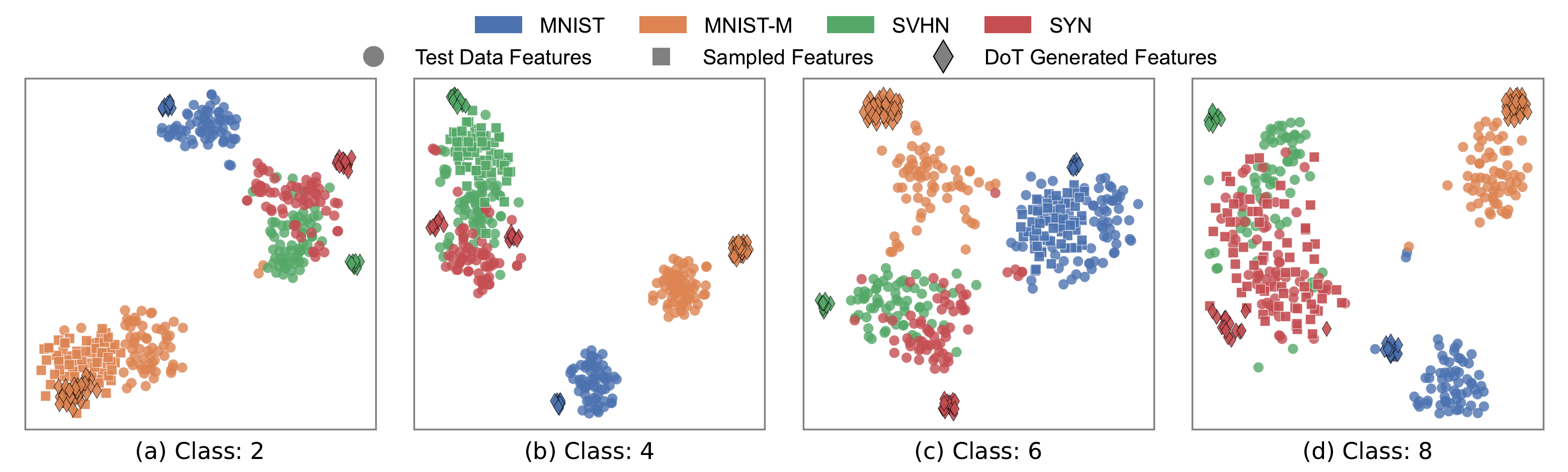}
    \vspace{-0.1cm}
\caption{t-SNE visualization of DoT features. 
We depict three kinds of final-layer features obtained from DoT-SLCA on Split DigitsDG, including
real features extracted from the test set of all domains (circle), 
in-domain features sampled from the preserved $\mathcal{H}_c$ (square, white edge),
and DoT's generated features of all domains (diamond, black edge). Points are colored according to domain identities. \textbf{a-d}, The features of class 2, 4, 6, and 8 from different tasks. Best viewed in color.}
\vspace{-0.2cm}
\label{fig.generated_feature}
\end{figure}

To further validate the efficacy of our approach, we use t-SNE to visualize the final-layer features of DoT-SLCA in Fig.~\ref{fig.generated_feature}. We compare three types of features: (1) real features extracted from the test set of all domains, (2) in-domain features sampled from the preserved $\mathcal{H}_c$, and (3) DoT's generated features of all domains.
It can be seen that the real features of all domains are naturally divided into multiple clusters, and the sampled in-domain features only align with one of them, suggesting the particular challenges of DGCL.
In contrast, the generated features exhibit appropriate alignment with the corresponding real-domain clusters, highlighting that the DoT module can effectively capture the domain-relevant information and combine it with semantic-relevant information, so as to address the DGCL problem.

\section{Conclusion}\label{sec.con}

In this work, we introduce domain generalizable continual learning (DGCL), a novel and realistic setting that addresses the unique challenges of domain generalization in continual learning. By performing an extensive empirical investigation of DGCL, we expose the severe limitations of state-of-the-art baselines in continually learning representations and aligning outputs for generalizing to unseen domains. To overcome these challenges, we propose an innovative approach that adaptively transforms semantic- and domain-relevant representations with pre-trained models, drawing inspiration from the robust mechanisms of the human brain. Our approach delivers significant improvements in DGCL, functioning as a plug-in strategy that supports both full parameter tuning and parameter-efficient tuning. 
These contributions not only advance the state-of-the-art in continual learning to accommodate unseen scenarios, but also set a promising direction for visual domain generalization in adapting to temporal-scale changes, with broad implications for deploying adaptive and robust AI systems in real-world applications where environments are both diverse and dynamic.

\section{Data Availability}
All the data used for this research is publicly accessible and proper citations are also provided in the article.

\section{Statements and Declarations}
The authors declare no competing interests.

\section{Acknowledgment} 
This work was supported by STI2030-Major Projects (2022ZD0204900), the NSFC Projects (Nos.~62406160, 32021002), Beijing Natural Science Foundation L247011, and the Tsinghua-Peking Joint Center for Life Sciences.

\bibliographystyle{ieeetr}
\bibliography{sn-bibliography}

\clearpage
\begin{appendices}




\end{appendices}


\end{document}